\documentclass{article}

\usepackage{arxiv}

\usepackage[utf8]{inputenc} 
\usepackage[T1]{fontenc}    
\usepackage{hyperref}       
\usepackage{url}            
\usepackage{booktabs}       
\usepackage{amsfonts}       
\usepackage{array, amsmath}
\usepackage{bm} 
\newcolumntype{M}{>{$}l<{$}}  
\usepackage{nicefrac}       
\usepackage{microtype}      
\usepackage{tabularx}
\usepackage{booktabs}
\usepackage{lipsum}
\usepackage{graphicx}
\usepackage{caption} 
\usepackage{longtable}
\usepackage{tikz}
\usepackage{float} 
\usepackage{authblk}
\usetikzlibrary{positioning,calc}
\usepackage{subcaption} 
\usepackage{wrapfig}  
\graphicspath{ {./images/} }
\usepackage{arydshln} 
\usepackage{colortbl} 
\DeclareMathOperator*{\argmin}{arg\,min}
\title{Inverse Reconstruction of Shock Time Series \\ from Shock Response Spectrum Curves\\ using Machine Learning}

\author{
Adam Watts\textsuperscript{*,\,1},
Andrew Jeon\textsuperscript{*, \,1}
Destry Newton\textsuperscript{*,\,1},
Ryan Bowering\textsuperscript{*,\,2},
 \\
\textsuperscript{1}Los Alamos National Laboratory, P.O. Box 1663, Los Alamos, NM 87545, USA \\
\textsuperscript{2}University of Rochester, Rochester, NY, USA \\
\texttt{\{acwatts, andrewjjeon, destryn\}@lanl.gov} \quad \texttt{rbowerin@ur.rochester.edu}
}

\begin{document}
\maketitle

\begingroup
  \renewcommand\thefootnote{\fnsymbol{footnote}}
  \footnotetext[1]{Alphabetical by first name, these authors contributed equally to this work.
  \hfill \break Code available at: \url{https://github.com/lanl/ML-Shock-Time-Series-Synthesis}}
\endgroup

\begin{abstract}
The shock response spectrum (SRS) is widely used to characterize the response of single-degree-of-freedom (SDOF) systems to transient accelerations. Because the mapping from acceleration time history to SRS is nonlinear and many-to-one, reconstructing time-domain signals from a target spectrum is inherently ill-posed. Conventional approaches address this problem through iterative optimization, typically representing signals as sums of exponentially decayed sinusoids, but these methods are computationally expensive and constrained by predefined basis functions.

We propose a conditional variational autoencoder (CVAE) that learns a data-driven inverse mapping from SRS to acceleration time series. Once trained, the model generates signals consistent with prescribed target spectra without requiring iterative optimization. Experiments demonstrate improved spectral fidelity relative to classical techniques, strong generalization to unseen spectra, and inference speeds three to six orders of magnitude faster. These results establish deep generative modeling as a scalable and efficient approach for inverse SRS reconstruction.
\end{abstract}


\section{Introduction}

A mechanical shock is a sudden, short-duration event characterized by a rapid change in velocity and typically accompanied by high acceleration. An event is considered a shock if its duration is significantly shorter than the natural period of the system under consideration \cite{lalanne_mechanical_2014}. The foundations of shock analysis were established by Biot, who introduced the study of single-degree-of-freedom (SDOF) systems subjected to transient dynamic loads, showing that the peak response of a system depends on its natural frequency and damping \cite{biot_transient_1932}. Building on this concept, the shock response spectrum (SRS) was developed to characterize shocks by recording the maximum response (displacement, velocity, or acceleration) of a set of SDOF systems, each with a different natural frequency, to a common input. This representation enables engineers to model and quantify the effects of shocks on linear systems across frequency ranges, regardless of the source. The SRS thus provides a standardized means of assessing shock severity, comparing shocks, and establishing equivalence criteria between measured transient environments and laboratory simulations \cite{european_cooperation_for_space_standardization_ecss_mechanical_2015}.

The SRS has become a standard tool across engineering disciplines for designing and qualifying assemblies and components to withstand transient shock loads. For example, MIL-STD-810G outlines environmental engineering considerations and laboratory test methods, including the use of SRS in equipment qualification \cite{united_states_department_of_defense_mil-std-810g_2008}. When plotted, the SRS displays natural frequency of an SDOF on the horizontal axis (with damping ratio specified by the user) and the peak response of that SDOF on the vertical axis. In this work, the shock input is taken to be a time series representing the acceleration applied to the system by the environment, with the response of interest being acceleration, denoted as $\ddot{x}(t)$. Of particular relevance is the \textit{maximax} spectrum, which records the maximum absolute acceleration response across all frequencies. However, the use of absolute values and maxima renders the maximax SRS both nonlinear and many-to-one, meaning that different acceleration time series can yield identical shock response spectra \cite{lalanne_mechanical_2014}.

In practice, environmental specifications (ES) for shock testing are derived from measured shock response spectra. At each frequency, a one-sided upper normal tolerance interval is calculated to represent the upper bound of observed responses, and repeating this process across frequencies produces a statistically conservative description of the environment \cite{meeker_statistical_2023}. These tolerance intervals form the foundation for defining shock testing requirements within the ES. Qualification testing then seeks to reproduce the specified shock environment --- often using a shaker table --- in order to verify system performance, durability, and functionality. However, because shaker control software requires inputs in the form of an acceleration time series rather than SRS curves, a fundamental difficulty arises: the SRS cannot be uniquely inverted to recover the underlying time series. This challenge is not limited to a nominal SRS, but applies even more strongly to statistical envelopes or tolerance bounds, which are abstractions of many different measured shocks. Such bounds describe only the maximum responses expected at each frequency, without preserving timing, phase, or correlation across frequencies. As a result, the underlying time series consistent with a statistical bound is not just unknown but may be physically unrealizable, making the generation of suitable test inputs an inherently approximate process \cite{lalanne_mechanical_2014}.

Existing reconstruction methods address this problem by casting it as an optimization task, typically representing the acceleration signal as a finite sum of exponentially decayed sinusoids. While effective in certain cases, such methods are computationally demanding, restricted by their choice of basis functions, and limited in their ability to reproduce complex shocks. This paper introduces a novel machine learning (ML) framework that directly learns the mapping from an SRS to an acceleration time series. Unlike traditional approaches, the proposed methodology does not presuppose basis functions, achieves substantially faster computation, and offers improved fidelity in reconstructing time series consistent with target or specification SRS curves.

\subsection{Contributions}
The main contributions of this work are summarized as follows:
\begin{itemize}
    \item \textbf{A conditional variational autoencoder (CVAE) framework for SRS inversion:}  
    We introduce a CVAE architecture that directly maps target shock response spectra (SRS) to corresponding time-domain acceleration signals, eliminating the need for explicit basis functions. 
    The trained model achieves high spectral fidelity while reducing inference time by several orders of magnitude compared to optimization-based synthesis methods such as SDS and SDS+GA.

    \item \textbf{A GPU-accelerated synthetic shock data generation system with an SRS operator:}
    We develop a PyTorch-based framework that generates synthetic shock data at scale and includes a fully batched, differentiable Shock Response Spectrum (SRS) operator. The system rapidly produces paired time-series signals and their corresponding SRS at arbitrary sampling rates and durations, enabling high-throughput model training, evaluation, and benchmarking.

    \item \textbf{Two complementary performance metrics:}  
    Because no single metric fully captures global spectral agreement, relative performance across individual realizations, and localized frequency-domain fidelity, we evaluate model performance using RMSLE (including a per-sample RMSLE accuracy statistic) and per-frequency dB error. 
    Together, these measures provide a unified evaluation framework that quantifies overall reconstruction quality, probabilistic case-by-case reliability, and localized spectral deviations.

    \item \textbf{A standardized benchmark dataset for reproducible evaluation:}  
    We release a comprehensive SRS–time series benchmark composed of four subsets:   
    (i) two datasets derived from \textit{operational shock test recordings} normalized to unit amplitude and anonymized to remove identifying test signatures ($\mathcal{D}_{\text{test}}^{\text{A}}$ and $\mathcal{D}_{\text{test}}^{\text{B}}$);  
    (ii) one dataset derived from earthquake recordings ($\mathcal{D}_{\text{test}}^{\text{C}}$ ); and  
    (iii) one dataset of synthetic shocks generated using the proposed framework, incorporating all four basis function types ($\mathcal{D}_{\text{test}}^{\text{D}}$).  
    These datasets establish a standardized and diverse reference for future research on shock synthesis and SRS inversion .
\end{itemize}

\section{Preliminaries and Problem Formulation} \label{sec:Prelims_Form}

\subsection{The Shock Response Spectrum (SRS)}
The SRS operator is a functional mapping from a time-domain acceleration signal to a vector of spectral magnitudes evaluated over a set of SDOF natural frequencies:
\[
\mathrm{SRS} : \ddot{x}(t) \longmapsto \bm{\mathfrak{s}}_{x},
\]
where the mapping is evaluated using a vector of SDOF evaluation natural frequencies, $\bm{\mathfrak{f}}$, and parameterized by the damping ratio, $\zeta$:
\[
\bm{\mathfrak{s}}_{x} = \mathrm{SRS}\!\big\{\ddot{x}(t), \bm{\mathfrak{f}}; \zeta \big\}.
\]
Here, $\ddot{x}(t)$ denotes the input acceleration time series, 
$\bm{\mathfrak{f}} = [\mathfrak{f}_{\text{min}}, \ldots, \mathfrak{f}_{\text{max}}]$ 
is the vector of SDOF evaluation natural frequencies, 
and $\zeta$ is the scalar damping ratio. 
Since $\zeta$ is held constant in this work, it is omitted hereafter for brevity. 
The Fraktur symbol $\mathfrak{f}$ is used to distinguish the evaluation natural frequencies of the \textit{response spectra} from the natural frequencies $f_n$ that appear in various trigonometric basis functions.

\begin{figure}[h!]
\centering                     
    \includegraphics[page=8, trim=0cm 0cm 0cm 0cm, clip, width=\linewidth]{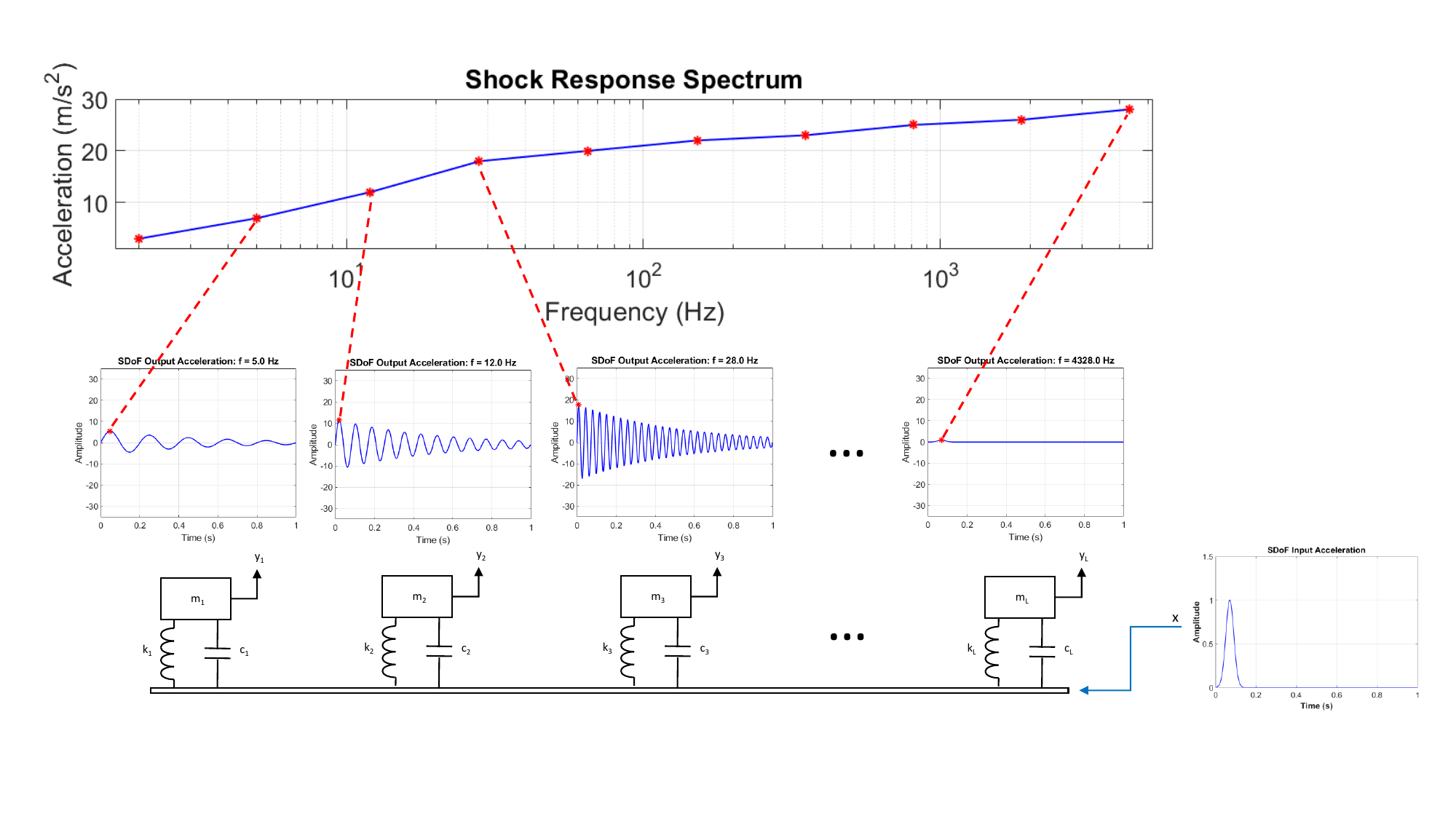}
\caption{Construction of the shock response spectrum (SRS) using a bank of single-degree-of-freedom (SDOF) oscillators. The top panel shows the SRS (blue) with selected evaluation frequencies highlighted (red markers). The middle row presents representative SDOF acceleration responses at those frequencies, illustrating frequency-dependent amplification and decay. The bottom schematic depicts the corresponding mass–spring–damper oscillators, each tuned to a natural frequency $\mathfrak{f}_i$ and subjected to a common base acceleration input (right), with each SRS ordinate obtained from the peak response of its oscillator.}
\label{fig:SRS}
\end{figure}

\subsubsection{Analytical Implementation of the SRS}

\begin{wrapfigure}{r}{0.45\textwidth}
    \vspace{-20pt} 
    \includegraphics[page=5, trim=9cm 2cm 7cm 3cm, clip, width=\linewidth]{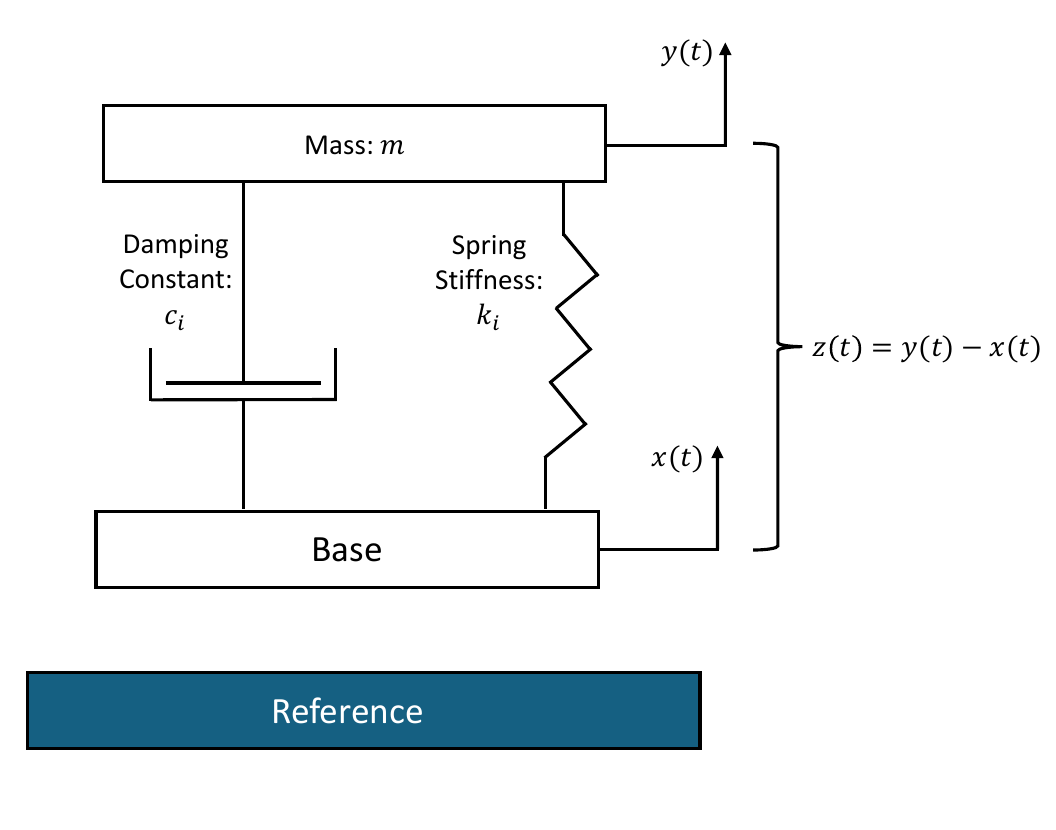}
    \vspace{-10pt}
    \caption{Base-excited SDOF oscillator with mass $m$, a particular spring stiffness $k_i$, and damping coefficient $c_i$ chosen such that the damping ratio $\zeta$ is constant across oscillators. The absolute displacement is $y(t)$, the base displacement is $x(t)$, and the relative displacement is $z(t) = y(t) - x(t)$.}
    \label{fig:sdof}
\end{wrapfigure}

The SRS provides a graphical representation of how a family of SDOF oscillators respond to a shock input time series. Conceptually, the SRS corresponds to the maximum response of uncoupled damped oscillators, each with a different natural frequency $\omega_i$, subjected to the same base excitation. The acceleration input at the base is denoted $\ddot{x}(t)$, and the absolute displacement of a given oscillator is $y(t)$. Figure~\ref{fig:sdof} illustrates such a system, consisting of a mass $m$, a frequency-dependent spring stiffness $k_i$, and a damping coefficient $c_i$ chosen so that the damping ratio $\zeta = c_i/(2m\omega_i)$ is fixed across oscillators, with relative displacement $z(t) = y(t) - x(t)$.

The \textit{maximax} acceleration SRS is defined as the maximum absolute acceleration response of the oscillator over time:  
\begin{equation}
\mathrm{SRS}\!\left\{\ddot{x}(t), \mathfrak{f}_i \right\} = \max_{t>0}\big(|\ddot{y}_i(t)|\big),
\label{eq:srs_def}
\end{equation}
where $\mathfrak{f}_i = \omega_i / (2\pi)$ is the natural frequency in Hz. 
Although the spectrum is formally defined through the oscillator response $\ddot{y}_i(t)$, 
it is conventional to write $\mathrm{SRS}\{\ddot{x}(t), \mathfrak{f}_i\}$ to highlight its dependence on the input shock $\ddot{x}(t)$. 
From equation \eqref{eq:srs_def}, it is evident that the SRS is not a one-to-one mapping: distinct acceleration time series may yield identical spectra.  

The relative displacement $z_i(t)$ of each oscillator evolves according to  
\begin{equation}
\ddot{z}_i(t) + 2\zeta\omega_i \dot{z}_i(t) + \omega_i^2 z_i(t) = -\ddot{x}(t),
\label{eq:sdof_eom}
\end{equation}
with $\omega_i = 2\pi \mathfrak{f}_i = \sqrt{k_i/m}$. Applying the Laplace transform gives  
\begin{equation}
\mathcal{L}\{z_i(t)\} = \int_{0}^{\infty} e^{-st} z_i(t) dt = Z_i(s),
\label{eq:laplace_def}
\end{equation}
and using the derivative property,  
\begin{equation}
\mathcal{L}\{\dot{z}_i(t)\} = s Z_i(s) - z_i(0).
\end{equation}
Substituting into equation \eqref{eq:sdof_eom} yields the system equation in the Laplace domain:  
\begin{equation}
(s^2 + 2\zeta\omega_i s + \omega_i^2)Z_i(s) - (s + 2\zeta\omega_i)z_i(0) - \dot{z}_i(0) = -\ddot{X}(s).
\label{eq:laplace_eom}
\end{equation}
Assuming zero initial conditions, which holds for SRS analysis, the equation simplifies, and the inverse Laplace transform gives the relative displacement:  
\begin{equation}
z_i(t) = -\frac{1}{\omega_d}\int_{0}^{t} \ddot{x}(\tau) e^{-\zeta\omega_i (t-\tau)} \sin\!\big(\omega_d(t-\tau)\big) \, d\tau,
\label{eq:zn_time}
\end{equation}
where $\omega_d = \omega_i \sqrt{1-\zeta^2}$ is the damped natural frequency. The absolute acceleration of the mass is then  
\begin{equation}
\ddot{y}_i(t) = \ddot{z}_i(t) + \ddot{x}(t).
\end{equation}

This analytical formulation forms the basis for the numerical implementation, where the same oscillator model is realized efficiently using digital filters.  

\subsubsection{Numerical Implementation of the SRS} 
In practice, the SRS of a discrete-time signal is computed using digital filters. Recent work has also developed differentiable implementations of IIR filters based on difference equations, enabling gradient-based optimization and machine learning integration \cite{defossez_hybrid_2022, yu_differentiable_2024}. Infinite impulse response (IIR) filters are particularly well suited due to their efficiency, and an implementation that we used in this work is available in the open-source \texttt{TorchAudio} library \cite{hwang_torchaudio_2023}.  The coefficients of these filters are standardized in ISO 18431-4 \cite{noauthor_mechanical_2007}, ensuring consistency across applications.  

This numerical approach is equivalent to simulating a bank of damped oscillators across the frequency range of interest. Each oscillator is defined by a mass $m$, a spring stiffness $k_i$ that sets its natural frequency
\[
\omega_i = \sqrt{k_i/m},
\]
and a damping coefficient $c_i$ chosen such that the damping ratio
\begin{equation}
\zeta = \frac{c_i}{2 m \omega_i}
\label{eq:zeta}
\end{equation}
remains constant across all oscillators. Thus, while the stiffness $k_i$ (and therefore the natural frequency $\omega_i$) varies from oscillator to oscillator, the damping coefficient $c_i$ is adjusted accordingly so that the damping ratio $\zeta$ remains fixed, as required in SRS analysis.

\subsection{Motivation for the Inverse SRS Problem}
\label{sec:inverse_motivation}
Throughout this paper, reconstructed acceleration time series are denoted simply as $\hat{x}(t)$, 
with the conventional double-dot notation, $\ddot{x}(t)$, suppressed for simplicity. In practical applications, engineers often operate in the \textit{response spectrum domain} rather than directly in the time domain. 
Measured shock events are represented as discrete-time acceleration realizations 
$\ddot{x}^{(j)}[n]$, $j=1,\ldots,J$, and each realization is mapped to its corresponding shock response spectrum:

\[
\boldsymbol{\mathfrak{s}}^{(j)}
=
\mathrm{SRS}\!\left\{
\ddot{x}^{(j)}[n],\, \boldsymbol{\mathfrak{f}}
\right\}
\in \mathbb{R}_{\ge0}^{F}.
\]

\begin{figure}[h!]
\centering                     
    \includegraphics[page=7, trim=0cm 0cm 0cm 0cm, clip, width=\linewidth]{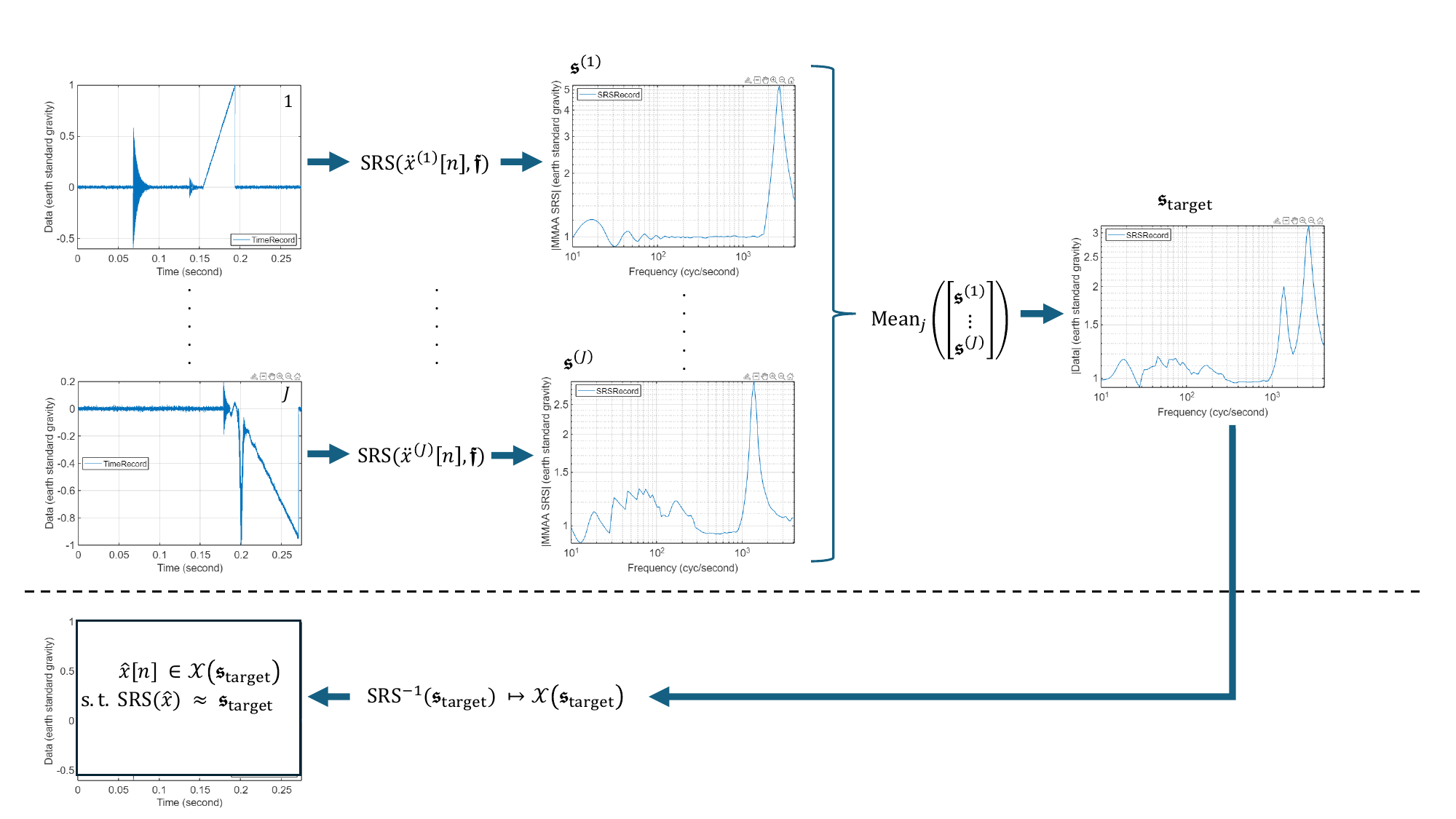}
\caption{
Forward and inverse formulations of the shock response spectrum (SRS) mapping.
Top: multiple acceleration time histories $\{x^{(j)}[n]\}_{j=1}^J$ are mapped through the SRS operator to corresponding spectral vectors $\{\bm{\mathfrak{s}}^{(j)}\}_{j=1}^J$, from which an example target spectrum $\bm{\mathfrak{s}}_{\mathrm{target}}$ is constructed (here shown as the ensemble mean). 
Bottom: the inverse SRS problem seeks admissible time-domain signals whose spectra match the target. 
Because the SRS operator is not injective, the inverse mapping is set-valued,
$\mathrm{SRS}^{-1}(\bm{\mathfrak{s}}_{\mathrm{target}}) \mapsto \mathcal{X}(\bm{\mathfrak{s}}_{\mathrm{target}})$,
where $\mathcal{X}(\bm{\mathfrak{s}}_{\mathrm{target}})$ denotes the set of acceleration time histories satisfying
$\mathrm{SRS}(\hat{x}) \approx \bm{\mathfrak{s}}_{\mathrm{target}}$.
The inverse problem therefore consists of selecting one admissible reconstruction $\hat{x}[n] \in \mathcal{X}(\bm{\mathfrak{s}}_{\mathrm{target}})$.
}
\label{fig:ProblemDescription}
\end{figure}

Performing the SRS operator on each acceleration realization produces a set of spectra 
$\{\boldsymbol{\mathfrak{s}}^{(1)}, \ldots, \boldsymbol{\mathfrak{s}}^{(J)}\}$, 
as illustrated in Fig.~\ref{fig:ProblemDescription}. 
Subsequent analyses, such as statistical aggregation or environmental specification (ES) development, 
are performed directly in this response spectrum space. Stacking the spectra across realizations yields,

\[
\boldsymbol{\mathfrak{S}}
=
\begin{bmatrix}
\boldsymbol{\mathfrak{s}}^{(1)} \\
\vdots \\
\boldsymbol{\mathfrak{s}}^{(J)}
\end{bmatrix}
\in \mathbb{R}^{J \times F}.
\]

Statistical aggregation is then performed along the realization dimension.
For example, the mean spectrum is defined as

\[
\boldsymbol{\mathfrak{s}}_{\mathrm{target}}
=
\frac{1}{J}
\sum_{j=1}^{J}
\boldsymbol{\mathfrak{s}}^{(j)}
=
\mathrm{Mean}_{j}\!\left(\boldsymbol{\mathfrak{S}}\right),
\]

which corresponds to an elementwise average across the evaluation natural frequencies. Similarly, an upper-tolerance spectrum is defined as,

\[
\mathfrak{s}_{\mathrm{target}}
=
\mathrm{UpperTol}\!\left(
\boldsymbol{\mathfrak{s}}^{(1)}, \ldots, \boldsymbol{ \mathfrak{s}}^{(J)}
\right) ,
\]

In both cases, $\boldsymbol{\mathfrak{s}}_{\mathrm{target}} \in \mathbb{R}_{\ge0}^{F}$ 
denotes a specification spectrum obtained through statistical operations performed in the SRS domain. Both operators are applied along the realization dimension and act elementwise with respect to frequency.

However, experimental reproduction of such an environment (e.g., via shaker-table qualification testing) requires an acceleration time history rather than a spectrum. 
Consequently, one must determine a discrete-time acceleration realization 
$\hat{x}[n]$ whose SRS matches the desired target spectrum:

\[
\mathrm{SRS}\!\left\{
\hat{x}[n],\, \boldsymbol{\mathfrak{f}}
\right\}
\approx
\boldsymbol{\mathfrak{s}}_{\mathrm{target}}.
\]

This requirement gives rise to the \textit{inverse SRS problem}: 
given a prescribed spectrum in the response domain, find a physically realizable acceleration history whose forward SRS mapping reproduces that spectrum as closely as possible.

\subsection{Inverse Problem Formulation} 
\label{sec:inverse_problem}

The shock response spectrum (SRS) operator maps a time-domain acceleration signal to a vector of spectral magnitudes evaluated over a set of single-degree-of-freedom (SDOF) natural frequencies:
\[
\mathrm{SRS} : \ddot{x}[n] \longmapsto \bm{\mathfrak{s}}_{x}.
\]
Given a measured or target SRS, denoted $\bm{\mathfrak{s}}_{\mathrm{target}}$, the goal of the inverse problem is to determine one or more acceleration time series $\hat{x}[n]$ whose response spectra match the target:
\[
\mathrm{SRS} : \hat{x}[n] \longmapsto \bm{\mathfrak{s}}_{\hat{x}} \approx \bm{\mathfrak{s}}_{\mathrm{target}}.
\]
In operator form, this inverse mapping can be expressed as
\[
\mathrm{SRS}^{-1} : \bm{\mathfrak{s}}_{\mathrm{target}} \longmapsto \hat{x}[n],
\]
where $\mathrm{SRS}^{-1}$ represents a generally non-unique inverse operator.

The SRS operator is \textit{surjective but not injective}: many distinct acceleration time series can produce identical response spectra, yet nearly any physically realizable spectrum corresponds to at least one admissible time-domain signal. Consequently, the mapping is not bijective, and an exact analytical inverse $\mathrm{SRS}^{-1}$ does not exist. The inverse SRS problem is therefore ill-posed and must be approximated by optimization or statistical learning methods that recover one of potentially many valid preimages (admissible time histories that yield the target spectrum).

Formally, the problem can be expressed as
\[
\hat{x}[n] =
\argmin_{x[n] \in \mathcal{X}}
\mathcal{L}_{\mathrm{SRS}}\!\left(
\bm{\mathfrak{s}}_{x}, \bm{\mathfrak{s}}_{\mathrm{target}}
\right),
\]
where $\mathcal{X}$ denotes the space of admissible acceleration signals (e.g., finite-length, square-integrable time series), and $\mathcal{L}_{\mathrm{SRS}}$ is a loss function comparing true and target spectra—commonly the root-mean-square logarithmic error (RMSLE) defined in Equation~\eqref{EQ:LossFunctionRMSE}. Traditional approaches, such as the sum of decayed sinusoids (SDS), solve this optimization explicitly. In contrast, the machine learning framework introduced in Section~\ref{sec:Methods} learns a statistical approximation of the inverse operator directly from data, enabling rapid generation of time histories consistent with target spectra without explicit optimization for each case.

\section{Related Work}
A common approach to reconstructing a candidate acceleration time series $\hat{x}(t)$ 
from a given SRS curve formulates the problem as an optimization task, 
typically using exponentially decayed sine functions as basis elements~\cite{smallwood_matching_1974}. 
A sum of exponentially decayed sinusoids can be expressed as:
\begin{equation}\label{EQ:SumDecayedSines}
\hat{x}_{\mathrm{SDS}}(t) = \sum_{i=1}^{M} A_{i} e^{-\lambda_{i} t} \sin{\!\left(2 \pi f_{i} t + \phi_{i}\right)},
\end{equation}
where the four parameters are the decay rate $(\lambda)$, phase $(\phi)$, amplitude $(A)$, 
and frequency $(f)$ in Hz. Although \eqref{EQ:SumDecayedSines} is written in continuous time, numerical implementation requires a discrete sequence $\hat{x}[n], \; n \in \mathbb{Z}$. Substituting $t = nT_s$, with sampling period $T_s = 1/f_s$, yields

\begin{equation}\label{EQ:SumDecayedSinesDiscrete}
\hat{x}_{\mathrm{SDS}}[n] = \sum_{i=1}^{M} A_{i} e^{\left( -\lambda_{i} n T_s \right)} \sin{ \left(2 \pi f_i n T_s + \phi_{i} \right)}.
\end{equation}

A typical loss function compares the target spectra to the reconstructed spectra 
using the root-mean-square logarithmic error (RMSLE), computed in base-10 logarithmic space:
\begin{equation}\label{EQ:LossFunctionRMSE}
\begin{aligned}
\mathfrak{s}_{\text{target},i} &= 
\mathrm{SRS}\!\left\{\ddot{x}_{\text{target}}[n], \mathfrak{f}_i \right\}, \\
\mathfrak{s}_{\text{SDS},i} &= 
\mathrm{SRS}\!\left\{\hat{x}_{\mathrm{SDS}}[n], \mathfrak{f}_i \right\}, \\
\mathcal{L}^{\text{SDS}}_{\text{SRS}} &=
\sqrt{\frac{1}{F}\sum_{i=1}^{F}
\left(
\log_{10}\!\left(
\frac{\mathfrak{s}_{\text{target},i}}
     {\mathfrak{s}_{\text{SDS},i}}
\right)
\right)^{2}}
\end{aligned}
\end{equation}

where \( F \) is the total number of evaluation natural frequencies from \( \bm{\mathfrak{f}} \),
and \( \mathfrak{f}_i \) denotes the \( i \)-th natural evaluation frequency sampled between 
\( \mathfrak{f}_{\min} \) and \( \mathfrak{f}_{\max} \).
For brevity, this base-10 logarithmic SRS loss is denoted simply as \(  \mathcal{L}^{\text{SDS}}_{\text{SRS}} \). The optimization problem can then be written as  

\begin{equation}\label{EQ:OptimizeSumDecayedSines}
\hat{\boldsymbol{\theta}}_{\text{SDS}} = 
\argmin_{\boldsymbol{\theta} \in \boldsymbol{\Theta}} \left( \mathcal{L}^{\text{SDS}} _{\text{SRS}} \right),
\end{equation}

where $\boldsymbol{\theta}$ is the vector of all parameters in equation \eqref{EQ:SumDecayedSinesDiscrete}.

Subsequent work has extended this approach in two main directions. One line incorporates temporal moments to better capture shock characteristics \cite{cap_methodology_1997}. Another broadens the basis set by introducing impulses and modified Morlet wavelets, linearly combined with the decayed sinusoids to increase representational flexibility \cite{brake_inverse_2011}.

Despite these advances, existing methodologies face several limitations. First, restricting the reconstruction to a finite set of basis functions yields shock signals that share similar forms and often lack realism. Ideally, a method should be basis-agnostic and able to span the full space of square-integrable functions, $L^2$. Second, because the mapping from SRS to time series is non-unique, current approaches return only a single reconstruction, which may not be feasible for a given shaker table. It would be advantageous to generate multiple distinct time series consistent with the same target spectrum. Third, the optimization is computationally intensive: using $M$ basis functions requires solving for $4M$ parameters, and the result is valid only for a specific SRS. Thus, current methods lack both speed and generalizability.

To the best of our knowledge, no machine learning approaches have yet been proposed in the literature to address these limitations.

\section{Notation}
This section summarizes the notation used throughout the paper. Unless otherwise stated, 
$t$ denotes continuous time, $n$ the discrete-time index, and $T_s = 1/f_s$ the sampling period. 

Table~\ref{tab:notation_mech} introduces the parameters of the base-excited single-degree-of-freedom (SDOF) oscillator, 
which forms the basis for defining the shock response spectrum (SRS). The corresponding discrete-time signal representation 
and SRS computation parameters are summarized in Table~\ref{tab:notation_srs}. 

The notation used for the classical sum-of-decayed-sinusoids (SDS) model and its loss function is given in 
Table~\ref{tab:notation_sds}, while the parameters and distributions employed to generate the synthetic dataset 
$\mathcal{D}_{\text{synth}}$ are described in Table~\ref{tab:DataGeneratorParams}. The datasets used in this study, 
including training, test, and auxiliary evaluation sets, are listed in Table~\ref{tab:datasets}. 

Finally, Table~\ref{tab:notation_ml} summarizes the machine learning notation, covering the latent-variable structure 
of variational autoencoders (VAEs) and conditional VAEs (CVAEs), the conditioning on SRS encodings, and the loss functions 
used for training. 

\section{Methods} \label{sec:Methods}


\begin{figure}[h!]
\centering
\includegraphics[
  page=1,
  trim=0 3.5cm 15cm 0,  
  clip,
  width=\linewidth,
  height=0.3 \textheight,
  keepaspectratio
]{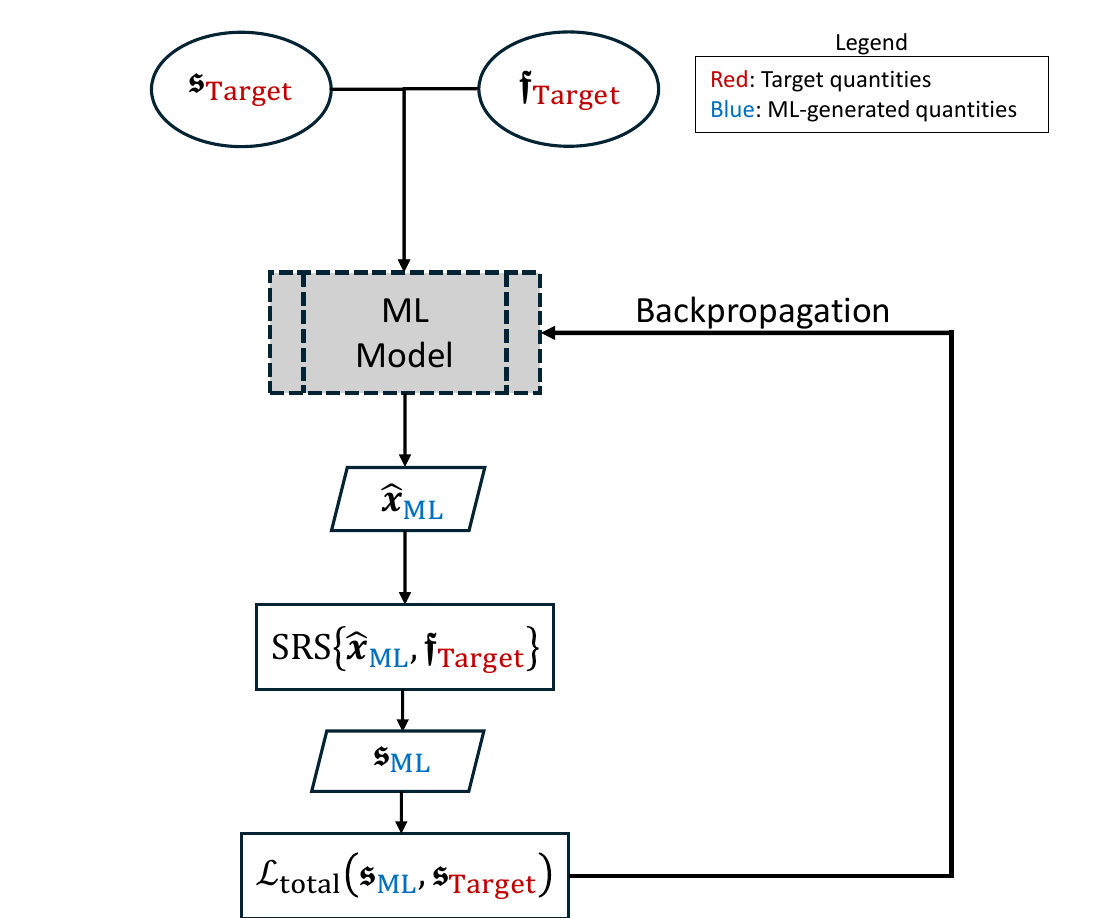}
\caption{Overview of the proposed machine-learning inverse SRS framework. The network receives the target SRS and corresponding natural frequency vector as inputs and produces a reconstructed acceleration time series. During training, the loss is computed between the SRS of the reconstructed signal and the target spectrum and is backpropagated to update model parameters. Additional loss terms, such as time-domain or regularization penalties, can be combined with the SRS loss to guide learning and improve stability.}
\label{fig:ml_inverse_srs_framework}
\end{figure}

\subsection{Possible ML Architectures}
As illustrated in Figure 2, the proposed inverse SRS framework replaces the generic “ML Model” block with different machine-learning architectures that map target SRS quantities to reconstructed time-domain signals and enable gradient-based optimization through a differentiable SRS operator. We began with a long short-term memory (LSTM) network \cite{hochreiter1997lstm}, but found that it consistently collapsed to producing nearly identical outputs across inputs, effectively predicting the mean. To address this, we experimented with transformers \cite{vaswani2023attentionneed}, motivated by their success in state-of-the-art generative tasks. However, in a regressive rather than autoregressive setting, transformers exhibited the same convergence behavior as LSTMs. Reformulating the task to predict parameters of a decayed sine model improved the realism of outputs, but still led to collapsed solutions.

We next turned to generative latent-variable models. A variational autoencoder (VAE) \cite{kingma_auto-encoding_2022} proved more effective, producing shocks with time-domain characteristics and SRSs similar to the target. To improve expressivity, we incorporated normalizing flows \cite{rezende2016variationalinferencenormalizingflows} into the latent space, which yielded only marginal improvements. More recent state-space sequence models were also considered, including Mamba \cite{gu_mamba_2024}, MinGRU \cite{fritschek2025mingrubasedencoderturboautoencoder}, and S4 \cite{DBLP:journals/corr/abs-2111-00396}. While promising in principle, these approaches either proved too slow for practical training or failed to learn meaningful dynamics in autoregressive mode, tending to collapse to trivial solutions when trained in parallel mode.

We are currently investigating latent diffusion and score-based diffusion methods, but chose to omit them for the scope of this initial work. We had hardware limitations that prevented extensive experimentation needed to develop and tune these models.

Given these challenges, we turned to a more structured generative approach: a conditional variational autoencoder (CVAE) tailored to the shock generation task. The CVAE provides a principled balance between modeling variability through a latent distribution and enforcing physical consistency via conditioning on the target shock response spectrum. In the following, we describe the architecture of the CVAE, including the encoder, decoder, and loss formulation.

The proposed CVAE is designed to generate time-domain shock signals conditioned on a target shock response spectrum (SRS). The model operates by concatenating each input time series, $\mathbf{x}$, with its corresponding SRS embedding, $\mathbf{s}$, thereby incorporating both temporal and spectral context. The encoder processes this joint representation through a sequence of one-dimensional convolutional layers with max-pooling and nonlinear activations, followed by flattening and fully connected layers. From the resulting hidden representation, two parallel linear layers output the mean and log-variance of a latent Gaussian distribution, enabling stochastic sampling via the reparameterization trick. Through empirical exploration, the latent dimension was set to 100. The decoder then combines the sampled latent vector with the SRS embedding and reconstructs a time-domain waveform. Training is guided by a composite objective that balances time-domain fidelity with frequency-domain consistency through the use of root-scaled mean log error (RSMLE) on the SRS and a custom waveform-shape loss, while a Kullback–Leibler divergence term regularizes the latent space. Details of the loss functions are provided in Section~\ref{sec:Loss}.

The encoder functions as a compression network, mapping temporal and spectral inputs into a compact latent representation that preserves the dynamic characteristics most relevant to the SRS. The decoder inverts this mapping, expanding the latent representation through fully connected layers, one-dimensional convolutions, and transposed convolutions that progressively upsample and refine the signal. A final convolutional layer reduces the representation to a single-channel time series. This architecture enables the CVAE to learn a structured latent space that captures variability in shock responses, while ensuring that generated signals remain consistent with the target SRS and exhibit physically plausible dynamic behavior. Section~\ref{sec:ML-VAE} provides additional details on this architecture.

\subsection{Variational Autoencoder (VAE)} \label{sec:ML-VAE}
Before introducing the Variational Autoencoder (VAE) framework, we clarify the notation for the signals of interest. 
In the mechanical SDOF model, the excitation is the base acceleration $\ddot{x}(t)$, from which the SRS is computed. In the machine learning context, we denote the corresponding discrete-time shock signal by the bold symbol $\mathbf{x}$, and reconstructed or generated shocks by $\hat{\mathbf{x}}$. 
This notation allows us to connect the physics-based model with the data-driven framework that follows.

Variational Autoencoders (VAEs) are a class of probabilistic models that learn a distribution over data \cite{kingma_auto-encoding_2022}. A VAE assumes that data $\mathbf{x}$ is generated through a latent variable model: first a latent $\mathbf{z}$ is drawn from a prior distribution $p(\mathbf{z})$, and then the observation $\mathbf{x}$ is generated from a conditional distribution $p_\theta(\mathbf{x}|\mathbf{z})$. These unobserved $\mathbf{z}$ are called \textit{latent variables}. The prior $p(\mathbf{z})$ is typically chosen as the isotropic Gaussian $\mathcal{N}(\mathbf{0}, \mathbf{I})$, which ensures a smooth and continuous latent space. The conditional distribution is parameterized by a flexible model $p_\theta(\mathbf{x}|\mathbf{z}) = f_\theta(\mathbf{z})$, where $f_\theta$ is a neural network (the decoder) with learnable parameters $\theta$. Figure \ref{fig:vae-inf-gen} provides an illustrative view of the VAE linking the encoder and deocder to their probabilistic definitions. 

The joint distribution over data and latent variables factorizes as
\[
p_\theta(\mathbf{x}, \mathbf{z}) = p_\theta(\mathbf{x}|\mathbf{z}) \, p(\mathbf{z}).
\]

The marginal likelihood of an observation is then
\[
\log p_\theta(\mathbf{x}) = \log \int p_\theta(\mathbf{x}|\mathbf{z}) \, p(\mathbf{z}) \, d\mathbf{z},
\]
but this integral is generally intractable when $f_\theta$ is highly nonlinear.

\paragraph{Variational approximation.} 
To deal with this intractability, an approximate posterior $q_\phi(\mathbf{z}|\mathbf{x}) = g_\phi(\mathbf{x})$ is introduced, parameterized by another neural network (the encoder). By multiplying and dividing inside the integral, we can rewrite the log marginal likelihood as
\[
\log p_\theta(\mathbf{x}) 
= \log \int q_\phi(\mathbf{z}|\mathbf{x}) \, \frac{p_\theta(\mathbf{x},\mathbf{z})}{q_\phi(\mathbf{z}|\mathbf{x})} \, d\mathbf{z},
\]
where $p_\theta(\mathbf{x},\mathbf{z}) = p_\theta(\mathbf{x}|\mathbf{z})p(\mathbf{z})$ is the joint distribution. 

Applying Jensen’s inequality yields a tractable lower bound, called the \textit{evidence lower bound} (or ELBO):
\begin{align}
\log p_\theta(\mathbf{x}) &\geq 
\mathbb{E}_{q_\phi(\mathbf{z}|\mathbf{x})}\!\left[ \log \frac{p_\theta(\mathbf{x},\mathbf{z})}{q_\phi(\mathbf{z}|\mathbf{x})} \right] \\
&= \mathbb{E}_{q_\phi(\mathbf{z}|\mathbf{x})}\!\left[ \log p_\theta(\mathbf{x}|\mathbf{z}) \right]
- D_{KL}\!\left(q_\phi(\mathbf{z}|\mathbf{x}) \,\|\, p(\mathbf{z})\right). \label{EQ:VLB}
\end{align}

\begin{figure}[h]
\centering
\begin{tikzpicture}[node distance=2.3cm, auto, >=stealth]

\node[draw, circle, minimum size=1cm] (x_inf) {$\mathbf{x}$};
\node[draw, circle, right=of x_inf, minimum size=1cm] (z_inf) {$\mathbf{z}$};
\draw[->] (x_inf) -- (z_inf);
\node[above=0.15cm of z_inf] {latent};
\node[above=0.15cm of x_inf] {observed};
\node[below=0.8cm of z_inf, align=center] {Approx.\ posterior \\ $q_\phi(\mathbf{z}|\mathbf{x})$};
\node[above=1.0cm of x_inf] {\textbf{Encoder:}  \text{Posterior inference} $q_{\phi}(z \mid x)$};

\node[draw, circle, minimum size=1cm, right=6.5cm of x_inf] (z_gen) {$\mathbf{z}$};
\node[draw, circle, right=of z_gen, minimum size=1cm] (x_gen) {$\mathbf{x}$};
\draw[->] (z_gen) -- (x_gen);
\node[above=0.15cm of z_gen] {latent};
\node[above=0.15cm of x_gen] {observed};
\node[below=0.8cm of z_gen, align=center] {Prior \\ $p(\mathbf{z})$};
\node[below=0.8cm of x_gen, align=center] {Likelihood \\ $p_\theta(\mathbf{x}|\mathbf{z})$};
\node[above=1.0cm of x_gen] {\textbf{Decoder:}  \text{Generative/Likelihood} $p_{\theta}(x \mid z)$ };

\node (midpoint) at ($(x_inf)!0.5!(x_gen)$) {};

\node[below=2.5cm of midpoint, align=center, text width=12cm] (eqs) {
\footnotesize
\textbf{Key relationships:} \\ 
$p_\theta(\mathbf{x},\mathbf{z}) = p_\theta(\mathbf{x}|\mathbf{z})\,p(\mathbf{z}), \quad
p_\theta(\mathbf{z}|\mathbf{x}) = \tfrac{p_\theta(\mathbf{x},\mathbf{z})}{p_\theta(\mathbf{x})}, \quad
p_\theta(\mathbf{x}) = \int p_\theta(\mathbf{x}|\mathbf{z})\,p(\mathbf{z})\,d\mathbf{z}$.
};

\end{tikzpicture}

\caption{Dual view of the Variational Autoencoder (VAE). 
Left: the \emph{inference} (encoder) direction $\mathbf{x}\!\to\!\mathbf{z}$ via $q_\phi(\mathbf{z}|\mathbf{x})$ that approximates the intractable posterior $p_\theta(\mathbf{z}|\mathbf{x})$. 
Right: the \emph{generative} (decoder) direction $\mathbf{z}\!\to\!\mathbf{x}$ via $p(\mathbf{z})$ and $p_\theta(\mathbf{x}|\mathbf{z})$. 
Maximizing the ELBO in Eq.~\ref{EQ:VLB} couples these two models.}
\label{fig:vae-inf-gen}
\end{figure}

The first term in \eqref{EQ:VLB} is a \textit{reconstruction term}, encouraging accurate recovery of $\mathbf{x}$ from $\mathbf{z}$, while the second term is a \textit{regularization term} that pushes the approximate posterior $q_\phi(\mathbf{z}|\mathbf{x})$ toward the prior $p(\mathbf{z})$.

\paragraph{Gaussian parameterization and reparameterization.} 
For tractability, both the prior and the approximate posterior are commonly chosen as Gaussian distributions:
\[
q_\phi(\mathbf{z}|\mathbf{x}) \sim \mathcal{N}(\boldsymbol{\mu}_q, \, \boldsymbol{\sigma}_q^2 \mathbf{I}), 
\quad p(\mathbf{z}) = \mathcal{N}(\mathbf{0},\mathbf{I}).
\]
In this case, the KL divergence in \eqref{EQ:VLB} has a closed form:
\begin{equation}\label{EQ:Simple_KL}
D_{KL}(q_\phi(\mathbf{z}|\mathbf{x}) \parallel p(\mathbf{z})) 
= \tfrac{1}{2}\sum_i \big(\boldsymbol{\mu}_{q,i}^2 + \boldsymbol{\sigma}_{q,i}^2 - \log \boldsymbol{\sigma}_{q,i}^2 - 1 \big).
\end{equation}

Since $\mathbf{z}$ is stochastic, gradients cannot flow directly through $q_\phi(\mathbf{z}|\mathbf{x})$. The reparameterization trick addresses this by expressing
\[
\mathbf{z} = \boldsymbol{\mu}_q + \boldsymbol{\sigma}_q \odot \boldsymbol{\epsilon}, 
\quad \boldsymbol{\epsilon} \sim \mathcal{N}(\mathbf{0}, \mathbf{I}),
\]
where $\odot$ is the elementwise product. This makes $\mathbf{z}$ a differentiable function of $(\mathbf{x}, \phi, \boldsymbol{\epsilon})$, enabling end-to-end training with backpropagation. 

\paragraph{Training and generation.} 
The encoder $g_\phi$ and decoder $f_\theta$ are trained jointly by maximizing the ELBO \eqref{EQ:VLB}. After training, new data samples can be generated by drawing $\mathbf{z}\sim p(\mathbf{z})$ and passing it through the decoder $f_\theta(\mathbf{z})$. 

\begin{figure}[h!]
\centering                     
    \includegraphics[page=4, trim=3.5cm 4cm 3.5cm 2cm, clip, width=0.5\linewidth]{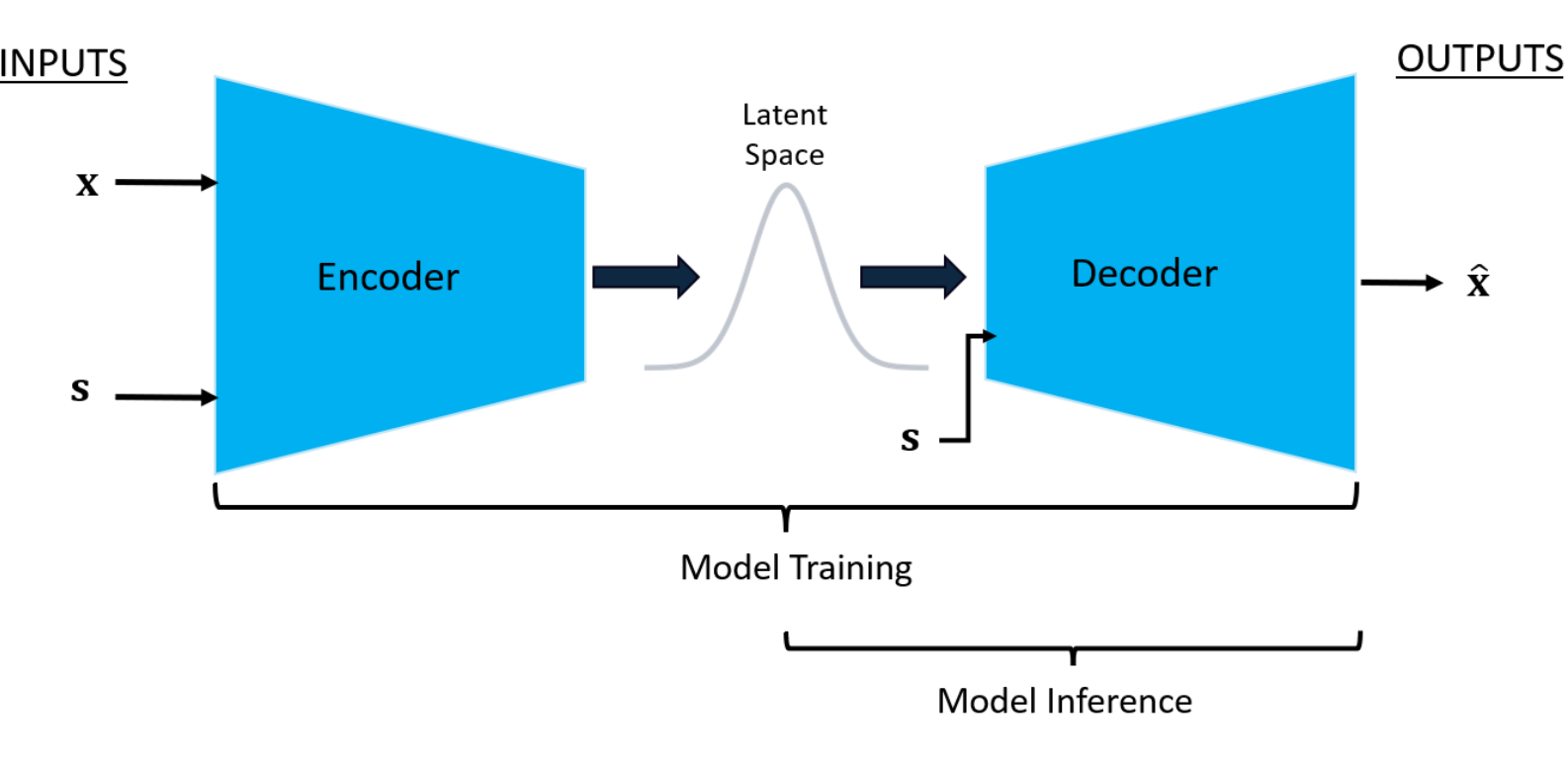}
\caption{Conditional VAE for shock generation. During \textbf{training}, the encoder ingests the shock signal $\mathbf{x}$ and the SRS encoding $\mathbf{s}$ to produce a latent representation. During \textbf{inference}, the decoder generates a shock $\hat{\mathbf{x}}$ conditioned on $\mathbf{s}$ (and a sampled latent).}
\label{fig:Encoder_Decoder}
\end{figure}

Since shocks need to be generated to match a given SRS, it is not sufficient to use a vanilla VAE for this task. Instead, a Conditional VAE (CVAE) \cite{sohn2015learning} is employed, in which an additional conditioning variable $\mathbf{s}$ is introduced. The context $\mathbf{s}$, which represents the encoding of the SRS at particular frequencies, is concatenated both to the input before encoding and to the latent vector before decoding, see Figure \ref{fig:Encoder_Decoder}.. This results in the encoder $q_\phi(\mathbf{z}\mid \mathbf{x},\mathbf{s})$ and decoder $p_\theta(\mathbf{x}\mid \mathbf{z},\mathbf{s})$. In practice, the conditional prior is usually chosen as $p(\mathbf{z}\mid \mathbf{s})=\mathcal{N}(0,\mathbf{I})$, i.e.\ independent of $\mathbf{s}$, while the conditioning enters through the encoder and decoder. This ensures that the generated shocks are consistent with the desired structural response defined by $\mathbf{s}$ while still being tractable. 

\subsection{Loss Function}\label{sec:Loss}
The CVAE is trained using four complementary loss terms. The goal of these losses is to capture both 
time-domain and frequency-domain discrepancies between the generated and reference shocks, while also 
regularizing the latent space.  

\paragraph{Error Definitions.}
We use the following error measures throughout this work. Let $\mathbf{x}\in\mathbb{R}^N$ denote the target time series and $\hat{\mathbf{x}}\in\mathbb{R}^N$ the corresponding prediction. All losses are computed per sample and then averaged over the batch. The \textit{mean-squared error (MSE)} is
\begin{equation}
\mathrm{MSE}(\mathbf{x},\hat{\mathbf{x}}) \;=\; \frac{1}{N}\sum_{i=1}^{N}\left(\hat{\mathbf{x}}_i - \mathbf{x}_i\right)^2 .
\end{equation}
The \textit{root-mean-squared error (RMSE)} is
\begin{equation}
\mathrm{RMSE}(\mathbf{x},\hat{\mathbf{x}}) \;=\; \sqrt{\mathrm{MSE}(\mathbf{x},\hat{\mathbf{x}})}.
\end{equation}
The \textit{mean-squared-log-error (MSLE)} computes the squared error in log space,
\begin{equation}
\mathrm{MSLE}(\mathbf{x},\hat{\mathbf{x}}) \;=\; \frac{1}{N}\sum_{i=1}^{N}\left(\log\!\left(\hat{\mathbf{x}}_i + \epsilon\right) - \log\!\left(\mathbf{x}_i + \epsilon\right)\right)^2 ,
\end{equation}
where $\epsilon > 0$ is a small constant used for numerical stability. The \textit{root-mean-squared-log-error (RMSLE)} is defined as
\begin{equation}
\mathrm{RMSLE}(\mathbf{x},\hat{\mathbf{x}}) \;=\; \sqrt{\mathrm{MSLE}(\mathbf{x},\hat{\mathbf{x}})}.
\end{equation}

\paragraph{Waveform-shape loss.}

The first term, $\mathcal{L}_{\text{shape}}$, measures differences in the full SDOF acceleration 
waveform shape around the maximum response peak. For each evaluation natural frequency
$\mathfrak{f}_i \in \{\mathfrak{f}_{\min}, \ldots, \mathfrak{f}_{\max}\}$, the SDOF acceleration waveform is computed across discrete time indices $n \in \{0,\ldots,N\}$. A total of 100 logarithmically spaced evaluation natural frequencies between 
$\mathfrak{f}_{\min}=10$~Hz and $\mathfrak{f}_{\max}=4096$~Hz were used for the SRS calculation.  

\begin{figure}[h!]
    \centering     
    \vspace{-10pt} 
    \includegraphics[page=1, trim=0cm 0cm 0cm 0cm, clip, width=0.5 \linewidth]{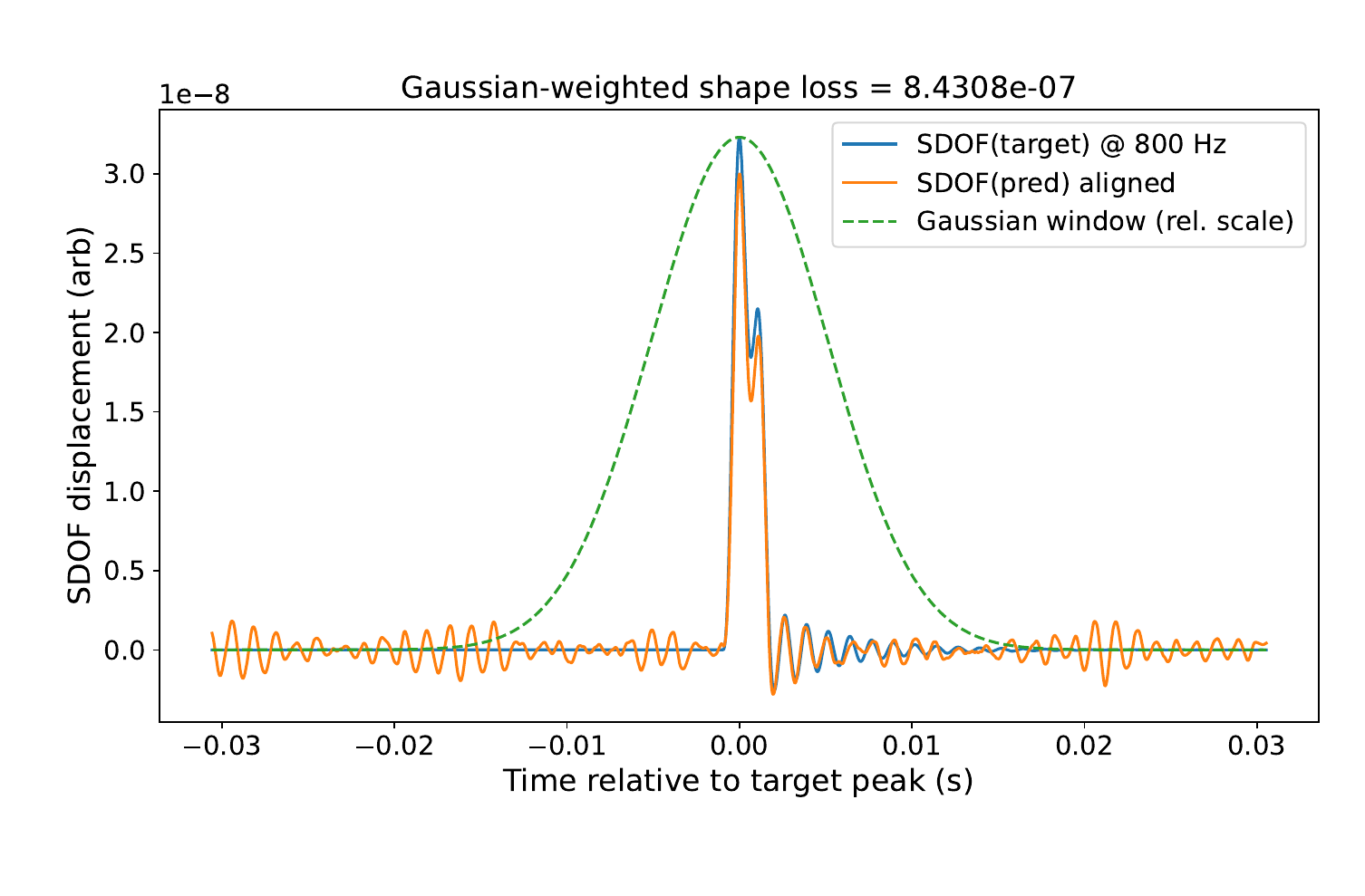}
    \caption{Example of the waveform-shape loss computed at a selected SDOF evaluation frequency. The predicted and target SDOF response signals are temporally aligned at their peaks, and a Gaussian weighting window emphasizes discrepancies near the primary shock feature. This localized loss formulation encourages accurate waveform reconstruction around the dominant response peak.}
    \label{fig:shape_loss}
\end{figure}

To align waveforms, the time index of the peak response is identified, and each waveform is shifted 
so its peak maximum is centered within a symmetric time window. A natural frequency-dependent Gaussian 
weighting, $w[n]$, is then applied, emphasizing regions near the maximum more strongly. See Figure \ref{fig:shape_loss} for an illustration of the waveform-shape loss at an example evaluation natural frequency of 800 Hz. The error is 
computed as the weighted sum of squared errors between the aligned SDOF waveforms:  
\begin{equation}
\text{SDOF}_{\text{shape}}(\mathbf{x}, \hat{\mathbf{x}})
= \tfrac{1}{N}\sum_{n=1}^{N} W_{\mathfrak{f}_i}[n] 
\Big(\, \hat{x}_{\mathfrak{f}_i}[n] - \tilde{x}_{\mathfrak{f}_i}[n] \,\Big)^2 
\in \mathbb{R}^{F},
\label{eq:srs_shape}
\end{equation}
where $\hat{x}_{\mathfrak{f}_i}[n]$ and $\tilde{x}_{\mathfrak{f}_i}[n]$ denote the aligned 
SDOF acceleration responses for the target and generated time series shocks, respectively.  

The scalar loss $\mathcal{L}_{\text{shape}}$ is then obtained by averaging across all natural frequencies:
\begin{equation}
\mathcal{L}_{\text{shape}}
= \tfrac{1}{F}
\sum_{\mathfrak{f}_i=\mathfrak{f}_{\min}}^{\mathfrak{f}_{\max}}
\text{SDOF}_{\text{shape}}(\mathbf{x}, \hat{\mathbf{x}}).
\label{eq:lshape}
\end{equation}

The batching notation is made explicit in Eq.~\eqref{eq:lshape} since this loss term has the most complex structure. For clarity and consistency, all other loss terms introduced in this section are defined analogously but are only implicitly averaged over the batch dimension.

If the Gaussian weighting were reduced to a Dirac delta function,
or equivalently to a zero standard deviation $\sigma \!\to\! 0$,
the shape loss would degenerate to comparing only the single
maximum SDOF response value at each evaluation frequency.
In this limiting case, $\mathcal{L}_{\text{shape}}$ effectively
recovers the conventional SRS definition based solely on the
maximum absolute response amplitude, thereby discarding all
temporal information surrounding the peak.
By instead adopting a finite Gaussian width $\sigma > 0$,
the weighting function $W_{\mathfrak{f}_i}[n]$ includes a small region around the maximum,
allowing nearby samples to contribute to the error computation.
This preserves local waveform morphology and yields smoother,
more informative gradients during optimization,
while still emphasizing the physically meaningful peak region of
the SDOF response.

\paragraph{Time-series loss.}
The second term, $\mathcal{L}_{\text{TS}}$, is the standard mean squared error between the reference 
and generated shock time series:  
\begin{equation}
\mathcal{L}_{\text{TS}} = \text{RMSE}(\mathbf{x}, \hat{\mathbf{x}}).
\label{eq:lts}
\end{equation}
This term corresponds to the negative log-likelihood term in the ELBO formulation when the variance is zero.

\paragraph{Frequency Domain Loss.}
We also evaluated frequency-domain objectives, including an FFT-based magnitude loss and a multi-scale STFT loss that compares spectro-temporal structure across multiple window sizes. The goal was to encourage both fine-grained spectral detail and longer-horizon temporal coherence. In practice, however, neither term produced consistent improvements in overall performance. Models trained with FFT or multi-scale STFT losses generally performed slightly worse than those trained with the simpler losses described above, although they produced marginally higher sample diversity at inference. In contrast, introducing a Welch power spectral density (PSD) loss improved the SRS accuracy of the generated time-series signals, and this Welch PSD term was ultimately adopted in our final training objective.

The Welch PSD loss, $\mathcal{L}_{\text{PSD}}$, is defined as the MSLE between the PSD of the ground-truth time series (from which the target SRS is computed) and the PSD of the generated time series. Here, $\mathbf{P}_{\mathbf{xx}}$ denotes the auto power spectral density (PSD) estimate of $\mathbf{x}$ computed via Welch's method.

\begin{equation}
\mathcal{L}_{\text{PSD}} = \text{MSLE}(\mathbf{P}_ \mathbf{\text{xx}}, \mathbf{P}_{\hat{\text{x}} \hat{\text{x}}} ).
\label{eq:lpsd}
\end{equation}

\paragraph{Shock Response Spectrum loss.}
The third term, $\mathcal{L}_{\text{SRS}}$, measures discrepancies between the target and predicted 
spectra. It is defined as the MSLE between the SRS values of the reference and 
generated shocks:
\begin{equation}
\mathcal{L}_{\text{SRS}} = \text{MSLE} \Big(\text{SRS}\{\mathbf{x}, \bm{\mathfrak{f}} \}, \,
\text{SRS}\{\hat{\mathbf{x}}, \bm{\mathfrak{f}} \}\Big).
\label{eq:lsrs}
\end{equation}

\paragraph{KL divergence.}
The final term is the standard KL divergence between the approximate posterior and the prior \ref{EQ:Simple_KL}, which 
encourages smoothness and continuity in the latent space.  

\paragraph{Total loss.}
The total loss function is a weighted combination of the five components,
\begin{equation}
\mathcal{L}_{\text{total}} =
 \lambda_{\text{shape}}\mathcal{L}_{\text{shape}}
+ \lambda_{\text{TS}}\mathcal{L}_{\text{TS}}
+ \lambda_{\text{PSD}}\mathcal{L}_{\text{PSD}}
+ \lambda_{\text{SRS}}\mathcal{L}_{\text{SRS}}
- \lambda_{\text{KL}} \tfrac{1}{2}\big(1 + \log(\boldsymbol{\sigma}_q^2) - \boldsymbol{\mu}_q^2 - \boldsymbol{\sigma}_q^2\big).
\label{eq:ltotal}
\end{equation}

We did some loss coefficient tuning in order to determine the most influential loss terms. Our most successful model had normalized loss coefficients of 0.404 KL Divergence Term, 0.282 Waveform Shape Loss, 0.237 log10 SRS Loss, 0.062 time-series loss, and 0.0147 Welch PSD Loss. As can be seen, the Waveform Shape Loss, KL Divergence and SRS loss were the dominant terms.

\subsection{SRS}
\subsubsection{Evaluation Natural Frequencies}
Shock time series are typically sampled at high rates to capture the high-frequency content characteristic of mechanical shocks and to minimize errors in calculating the SRS. Inadequate sampling can introduce significant errors when the maximum SRS frequency approaches the Nyquist limit of the sampled signal, $2/T_s$.  

For a sinusoidal response under zero damping, the worst-case percentage error in estimating the true peak SRS response is given by
\begin{equation}\label{EQ:SRS_error}
\mathrm{e}_{\mathrm{SRS}} = 100 \left(1 - \cos\!\left(\frac{\pi}{S_f}\right)\right),
\end{equation}
where \(S_f = \tfrac{f_s}{\mathfrak{f}_{\max}}\) is the ratio of the sampling rate, \(f_s\), to the maximum SRS frequency, \(\mathfrak{f}_{\max}\) \cite{lalanne_mechanical_2014}. This error arises because the true oscillator peak may not coincide with the discrete sampling instants.  

In practice, most of this error can be mitigated by \textit{up-sampling} (interpolating to a higher rate) prior to computing the SRS. For example, up-sampling a 32.768~kHz signal by a factor of six reduces the maximum error to less than 0.25\%, which is negligible for the confidence and coverage levels typically required in qualification testing.  

For large-scale training, however, repeated up-sampling was computationally prohibitive since thousands of SRS evaluations were required. To balance efficiency with accuracy, we limited the maximum SRS frequency to \(\mathfrak{f}_{\max} = 4096\)~Hz, corresponding to one-eighth of the 32.768~kHz sampling rate. Under this configuration, Eq.~\eqref{EQ:SRS_error} bounds the maximum error across the frequency domain at 7.61\%, which was deemed acceptable for this study.  

The set of evaluation natural frequenciess, $\bm{\mathfrak{f}}$, was chosen as 100 logarithmically spaced samples between \(\mathfrak{f}_{\min}=10\) and \(\mathfrak{f}_{\max}=4096\)~Hz. Alternative schemes, including linear spacing and random sampling within this range, were also tested. Logarithmic spacing provided the most effective trade-off between model accuracy and training efficiency, and was therefore adopted in all experiments.

\subsubsection{Shock Padding}
Because the maximax response may occur after the shock ends, the time series must be padded with zeros during the calculation \cite{lalanne_mechanical_2014}. Since the amplitude of the response decays after the shock, padding with half a cycle of the minimum natural frequency of interest is sufficient. The padding length is given by  
\begin{equation}
L = \left\lceil \frac{f_s}{2 \, \mathfrak{f}_{\min} \sqrt{1-\zeta^2}} \right\rceil,
\label{eq:padding}
\end{equation}
where $f_s$ is the sampling frequency and $\lceil \cdot \rceil$ denotes the ceiling operator.

Calculating the SRS requires padding the end of the shock signal with zeros according to \eqref{eq:padding}. 
With larger batch sizes or a dense frequency grid, this zero-padding can make SRS computations computationally expensive. 
To reduce cost, we investigated scaling the padding length to $\lceil \tfrac{L}{p}\rceil $, where $L$ is the full padding length and $p$ is a scale factor. For our data using a sample rate of $f_s=32768$ Hz, $\mathfrak{f}_\text{min}=10$ Hz, and damping factor of $\zeta=0.03$, this yields a full padding length of $L=1640$ elements. 

To evaluate this approach, 10{,}000 synthetic shocks were generated and their SRS values were first computed with full padding. 
The SRS was then recomputed with 100 evenly spaced values of $p \in [2,4]$, and the mean absolute error (MAE) between each scaled-padding SRS and the fully padded SRS was recorded. 
Across this range, the MAE remained extremely small: below $1\times 10^{-4}$ for both shocks and pure noise up to $p=3$. 
Beyond this point, the error increased more noticeably, reaching approximately $7\times 10^{-4}$ for shocks and $1.1\times 10^{-4}$ for noise at $p=4$.

Because untrained models initially produce random outputs, we also examined the effect of padding on pure noise. 
Specifically, random signals drawn from $\mathcal{N}(0, 0.3^2)$ were generated, and their SRS errors were measured in the same way. 

Based on the experiments above, a padding scale of $p=3$ was adopted for all SRS calculations in this work, thereby reducing the padding length from 1640 elements down to 547 elements. At this level, the MAE remains below $1\times 10^{-4}$, ensuring negligible loss of accuracy while reducing computational padding cost by a factor of three.


\subsection{Encoding}
The SRS of a shock signal is computed by evaluating the acceleration response at a set of natural frequencies. The choice of these frequencies can strongly influence model performance. Because the raw SRS values alone do not contain information about the frequencies at which they are measured, it is necessary to embed frequency information into the representation before passing it to the model. 

To achieve this, the vector of evaluation natural frequencies, $\bm{\mathfrak{f}}$, was used directly as a positional encoding. 
The raw SRS vector, $\bm{\mathfrak{s}}$, were reweighted by their corresponding natural frequencies and then log-scaled according to
\begin{equation}\label{EQ:Freq_Encoding}
\bf{s} = \log_{10}\!\big(\bm{\mathfrak{s}} \odot \bm{\mathfrak{f}}\big),
\end{equation}
where $\odot$ denotes elementwise multiplication.

\subsection{Data}
One significant challenge in this domain is the absence of a well-established, widely recognized dataset of time series and their corresponding SRS curves for training and evaluating inverse SRS algorithms. By contrast, in the image processing field, canonical datasets like the MNIST database of handwritten digits provide a standardized benchmark for comparing and refining machine learning methods. To address this gap, the following subsections outline a process for generating time series data and their SRSs, designed to create a rich and diverse dataset of synthetic shock time series, offering high variability and complexity.
\subsubsection{Synthetic Shock Generation}
\begin{figure}[h!]
    \centering
    \includegraphics[page=1, trim=0cm 0cm 0cm 0cm, clip, width=\linewidth]{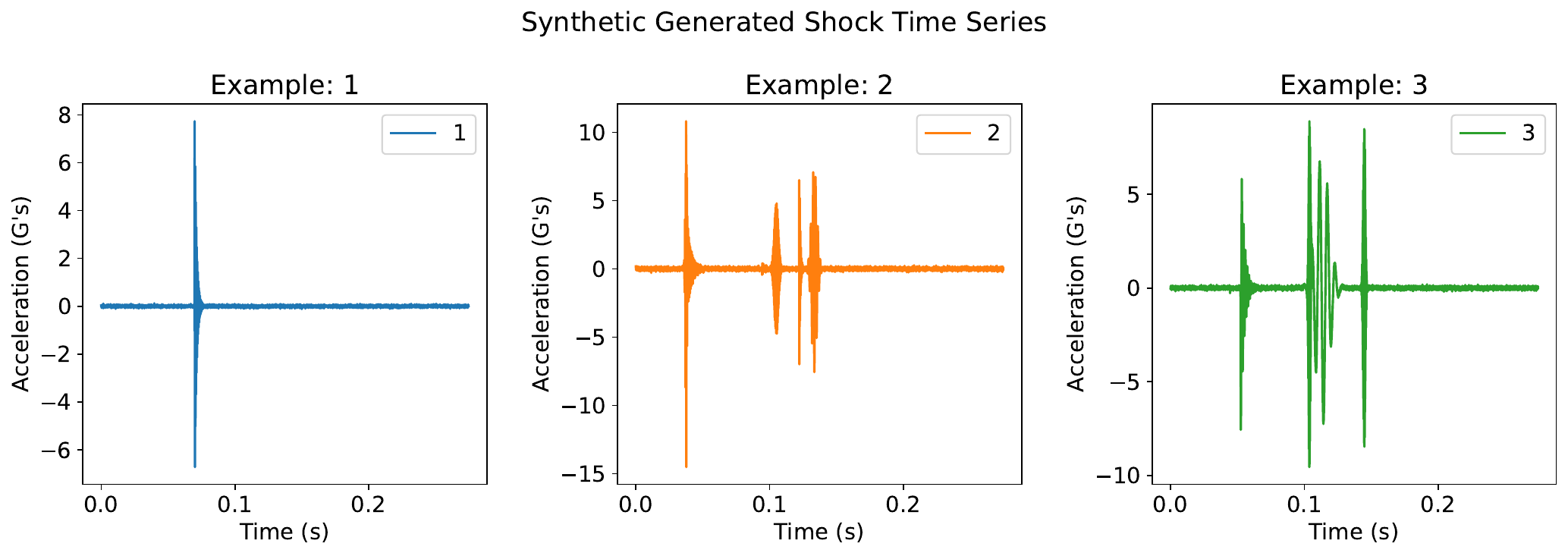}
    \caption{Examples of three different synthetic shock time series generated by the data generation process.}
    \label{fig:synthetic_shocks}
\end{figure}

The synthetic data generation framework employs two distinct basis functions:  
(1) an exponentially decaying sinusoid and  
(2) an exponential Morlet-like pulse. 
While earlier concerns in the literature have focused on basis dependence in the \textit{inverse SRS problem}, here the basis functions are not used for reconstruction but rather for producing large datasets of physically realistic shocks for machine learning. We found that these two basis functions produced the most effective synthetic dataset, as models trained solely on them generalized well to real shock data, even without exposure to real shocks during training. We initially experimented with additional bases, including radial basis functions (RBFs) and sawtooth functions, but these did not generalize as effectively as the first two. Our data generation code can generate shocks utilizing any combination of the four.

The synthetic shock time series are generated at the same sampling rate as the field data, $f_s = 2^{15} = 32.768$ kHz, with a fixed length of $N = 9000$ samples (corresponding to $\approx 0.2747$ s). This duration ensures coverage of spectral content down to at least 10 Hz while accommodating the high sampling rate: three full cycles of a 10 Hz wave are present, providing sufficient resolution for spectral analysis. The minimum natural frequency for SRS evaluation was therefore set to $\mathfrak{f}_{\text{min}} = 10$ Hz. 

Each synthetic shock is generated by summing $\mathcal{B}$ randomly chosen basis functions $\psi_k[n]$, where $k \in \{1,2\}$ is sampled uniformly. The basis functions are defined as
\begin{equation}\label{EQ:BasisCases}
\psi_k[n] =
\begin{cases}
A_i \exp\!\left(-\lambda_i n T_s \right) 
   \sin\!\left( 2\pi f_i n T_s + \phi_i \right), 
   & k = 1, \\[1.2em]
A_i \exp\!\Big[ \,\eta_i \omega_i \,\big(\ln(1+nT_s) - nT_s \big)\,\Big] 
   \cos\!\left(\omega_i n T_s + \phi_i \right),
   & k = 2,
\end{cases}
\end{equation}
where $A_i, f_i, \lambda_i, \phi_i, \text{and} \ \eta_i$ are randomly sampled parameters (see Table~\ref{tab:DataGeneratorParams}).

To increase variability, each basis function is randomly placed within the time series. 
The placement is determined by the padding fraction $\xi_i \sim \mathcal{U}(0,0.75)$, 
which specifies the fractional offset relative to the total length $N$. 
Thus, $\xi_i=0$ corresponds to a basis starting at the very beginning of the time series, 
while $\xi_i=0.75$ places the basis as late as 75\% of the total length. 
The start index for the $i$-th basis is then given by $n \geq \lceil \xi_i N \rceil$. The full synthetic time series is then expressed as

\begin{equation}\label{EQ:SyntheticData}
\ddot{x}_{\text{synth}}[n] = \epsilon + \sum_{i=1}^{\mathcal{B}}
\begin{cases} 
0, & \text{if } n  <  \lceil \xi_{i} N \rceil, \\[0.8em]
\psi_k[n], & \text{else},
\end{cases}
\quad : \quad \epsilon \sim  \mathcal{N}(0, \sigma_{\epsilon}^2).
\end{equation}

When $\mathcal{B}$, each subsequent basis function has an equal probability $p=0.5$ of adopting the same padding fraction as the previous basis. This is governed by the Bernoulli parameter $\gamma$:
\begin{equation}\label{EQ:PaddingAdoption}
\xi_{i+1} =
\begin{cases} 
\xi_{i}, & \text{if } \gamma = 1, \\
\xi_{i+1}, & \text{else},
\end{cases}
\quad \gamma \sim \text{Bernoulli}(p=0.5).
\end{equation}

This Boolean mechanism produces more intricate temporal patterns by occasionally aligning basis functions in time.

Finally, stationary Gaussian noise $\epsilon$ is applied to emulate environmental and measurement variability. 
The variance parameter $\sigma_{\epsilon}^2$ is sampled once from its uniform distribution 
(Table~\ref{tab:DataGeneratorParams}) and then held fixed for all generated shocks, 
ensuring a consistent level of background variability across the dataset. 
The distributions of all parameters are summarized in Table~\ref{tab:DataGeneratorParams}. 

\subsubsection{Training Data}
For model training, we generated a synthetic dataset of $d_\text{synth}=400{,}000$ time series using Eq.~\ref{EQ:SyntheticData}. 
This dataset is denoted $\mathcal{D}_{\text{synth}} \in \mathbb{R}^{d_\text{synth} \times N}$, where $N=9000$ is the signal length. 
In addition, we used a real shock dataset $\mathcal{D}_{\text{real}} \in \mathbb{R}^{d_\text{real} \times N}$ with $d_\text{real}=99{,}682$. 
The combined dataset is defined as $\mathcal{D}_{\text{combined}} = \mathcal{D}_{\text{synth}} \cup \mathcal{D}_{\text{real}}$, 
resulting in $d_\text{combined}=499{,}962$ shock time series. 
We applied a train--test split of 99\% and 1\%, yielding 
$\mathcal{D}_{\text{train}} \in \mathbb{R}^{d_\text{train} \times N}$ with $d_\text{train}=494{,}685$ 
and 
$\mathcal{D}_{\text{test}} \in \mathbb{R}^{d_\text{test} \times N}$ with $|\mathcal{D}_{\text{test}}|=d_\text{test}=4{,}997$. 
Here $|\mathcal{D}|$ denotes the cardinality (number of time series) in dataset $\mathcal{D}$.

\subsubsection{Normalization}
To ensure numerical stability and consistent input scaling across various datasets, 
each shock response spectrum (SRS) in the training set is normalized using a max-norm scheme such that its maximum value equals one, 
bounding the SRS within the range $[0,\,1]$. 
Because each training sample consists of a paired time-domain acceleration signal and its corresponding SRS, 
the associated time series is scaled by the same normalization factor, thereby preserving the linear relationship between the SRS and its generating waveform.

Since the normalization is defined by the SRS magnitude rather than by the peak amplitude of the time-domain signal, 
the normalized acceleration waveform is not constrained to lie within $[-1,\,1]$ and may exceed unity in magnitude when the peak time-domain response exceeds the maximum SRS value. 
As both the acceleration signal and its SRS are linearly scalable, this normalization procedure does not alter the underlying physical characteristics of the shock or its relative spectral content.

During inference, only the target SRS is available. 
Accordingly, the input SRS is normalized using the same max-norm convention applied during training and provided to the trained model to generate a normalized time-series realization. 
This generated time series is then rescaled by the reciprocal of the SRS normalization factor to recover the acceleration signal in physical units.

\begin{figure}[H] 
    \begin{subfigure}{0.49\textwidth}
        \centering
        \includegraphics[width=1.0\textwidth]{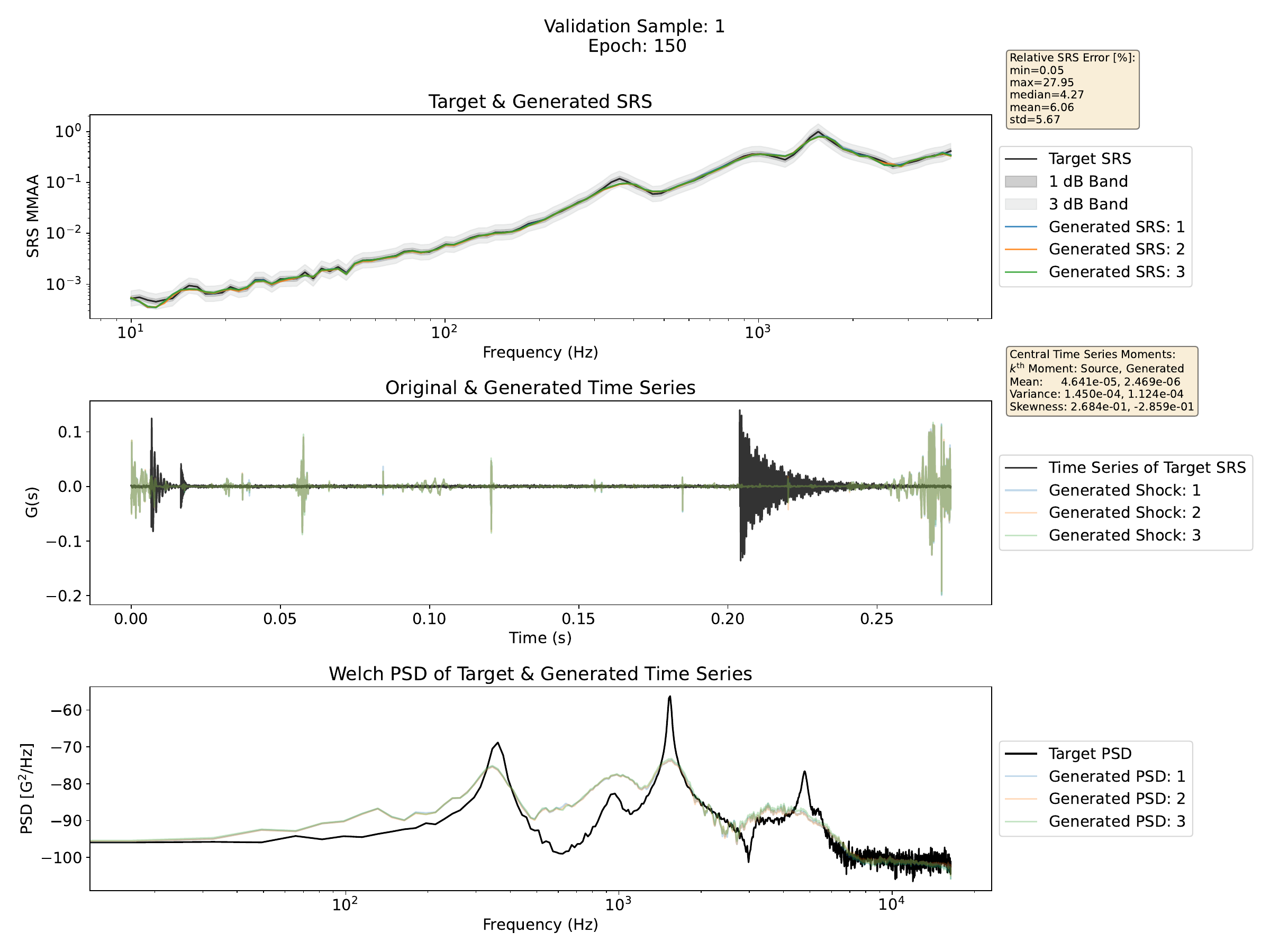}
    \end{subfigure}
    \begin{subfigure}{0.49\textwidth}
        \centering
        \includegraphics[width=1.0\textwidth]{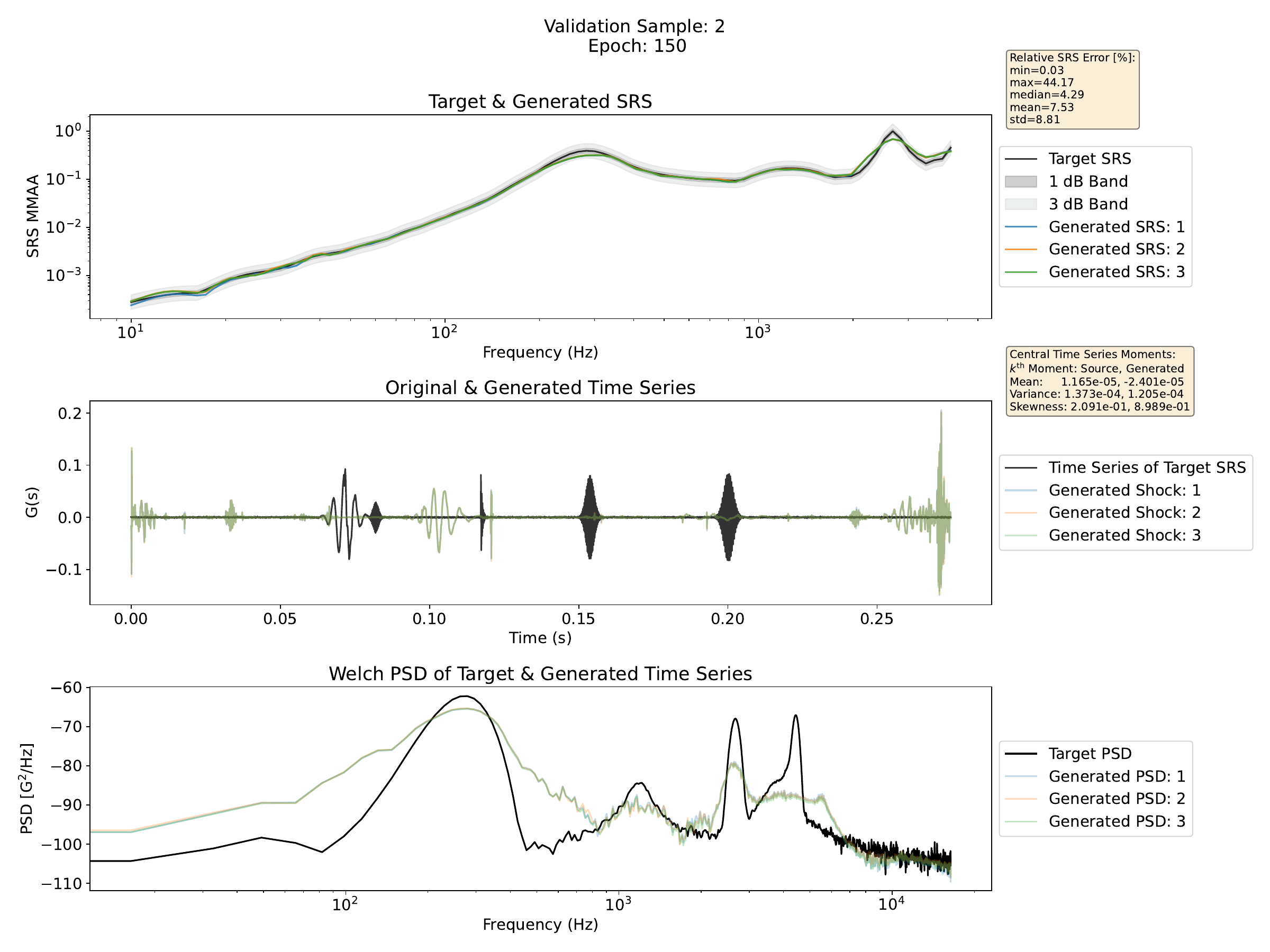} 
    \end{subfigure}
    \caption{Two representative elements from the test dataset $\mathcal{D}_{\text{test}}$, sample one is shown on the left and sample two is on the right. 
    The target SRS curves, $\bm{\mathfrak{s}}_{\mathrm{target}}^{1}$ and $\bm{\mathfrak{s}}_{\mathrm{target}}^{2}$, and the reconstructed SRS curves, 
    $\bm{\mathfrak{s}}_{\mathrm{ML}}^{1}$ and $\bm{\mathfrak{s}}_{\mathrm{ML}}^{2}$, are shown at the top subplot. 
    The target SRS curves shown on the top subplot originate from time series, $\ddot{x}^{(1)}$ and $\ddot{x}^{(2)}$, shown in the middle subplot. Similarly, the ML SRS curves originate from time series generated by our CVAE ML model, $\hat{x}_{\mathrm{ML}}^{(1)}$ and $\hat{x}_{\mathrm{ML}}^{(2)}$, are also shown in the middle subplot. Three generative time series samples are drawn from the latent space, denoted by different colors. For example, three realizations of $\hat{x}_{\mathrm{ML}}^{(1)}$ are shown.
    The Welch PSD of the corresponding target time series and ML time series realizations are also shown. }
    \label{fig:logspace_test_samples}
\end{figure}
\section{Results}\label{sec:Results}

A conditional variational autoencoder (CVAE) with 100 latent variables was trained on a combined dataset of 
$\mathcal{D}_{\text{synth}}$ ($d_{\text{synth}}=400{,}000$ synthetic shocks) and 
$\mathcal{D}_{\text{real}}$ ($d_{\text{real}}=99{,}682$ real shocks), 
resulting in a total of $d_{\text{combined}}=499{,}682$ shock time series. 
Training was performed for 150 epochs. 
For evaluation, the shock response spectrum (SRS) was computed at $F=100$ log-spaced natural frequencies 
ranging from $\mathfrak{f}_{\min}=10$~Hz to $\mathfrak{f}_{\max}=4096$~Hz. 

Figure~\ref{fig:logspace_test_samples} illustrates two representative samples from the test set $\mathcal{D}_{\text{test}}$, demonstrating agreement between the target and generated SRS curves, along with their corresponding time-series realizations and Welch PSD spectra.
In both examples, the generated responses exhibit close agreement with the targets across all domains.
The SRS curves are nearly indistinguishable over most of the frequency range, indicating that the CVAE accurately preserves the underlying shock energy distribution.
Likewise, the reconstructed time histories capture the overall waveform shape and transient features of the original signals.
The Welch PSD provides additional confirmation in the frequency domain, with the generated responses closely matching the broadband structure and dominant frequency content of the targets. The Welch PSD loss plot matches the ground truth closely up to approximately  $\mathfrak{f}_{\max}=4096$~Hz because this was the maximum natural frequency used by the SRS, whereas the maximum frequency used by Welch's method is the Nyquist frequency $(\frac{f_s}{2}$ Hz).
Taken together, these results highlight the ability of the CVAE to generate physically consistent, high-fidelity shock signals that reproduce global spectral characteristics while maintaining detailed agreement in both the time and frequency domains.

\subsection{Evaluation on Held-Out Datasets}\label{sec:Held-Out Datasets}

We evaluated the performance of the trained CVAE model across four distinct hold-out subsets, denoted 
$\mathcal{D}_{\text{test}}^{j}$ where $j \in \{A, B, C, D\}$. These are auxiliary held-out datasets for evaluation, not part of original train/test split.
Datasets $\mathcal{D}_{\text{test}}^{\text{A}}$ and $\mathcal{D}_{\text{test}}^{\text{B}}$ correspond to operational shock test recordings that retain their original physical amplitude scaling. 
Figure~\ref{fig:sds_vs_ML} illustrates an example target time series $(\ddot{x}_{\text{target}}[n])$ and its corresponding SRS $(\mathfrak{s}_{\text{target}})$ drawn from $\mathcal{D}_{\text{test}}^{\text{A}}$. 

These datasets are not part of the public release due to the inclusion of identifying test signatures; however, normalized and anonymized versions of $\mathcal{D}_{\text{test}}^{\text{A}}$ and $\mathcal{D}_{\text{test}}^{\text{B}}$ are included within the released benchmark.
For completeness, model evaluations were also performed on the original, non-normalized versions of 
$\mathcal{D}_{\text{test}}^{\text{A–B}}$ to assess model behavior under true physical amplitude scaling. 

Dataset $\mathcal{D}_{\text{test}}^{\text{C}}$ is derived from earthquake recordings whose original signals spanned 9–41~s at lower sampling rates. Finally, $\mathcal{D}_{\text{test}}^{\text{D}}$ consists of synthetic shock time series generated using our data-generation framework, encompassing all four basis function types. 

All time series in the held-out datasets were therefore either natively sampled or resampled to a uniform rate of $32.768$~kHz. 
Time series' requiring subsetting were truncated to a fixed duration of $\approx 0.2747$~s, with the maximum amplitude positioned at approximately $5\%$ of the time window.
$\mathcal{D}_{\text{test}}^{\text{A}}$, $\mathcal{D}_{\text{test}}^{\text{B}}$, and  $\mathcal{D}_{\text{test}}^{\text{D}}$, were natively sampled at $32.768$~kHz while $\mathcal{D}_{\text{test}}^{\text{C}}$ was resampled to $32.768$~kHz.
The data sets $\mathcal{D}_{\text{test}}^{\text{A}}$, $\mathcal{D}_{\text{test}}^{\text{B}}$, and $\mathcal{D}_{\text{test}}^{\text{C}}$ required  temporal subsetting to match a consistent $\approx 0.2747$~s duration. 

These preprocessing steps ensured consistent temporal resolution and signal length across all datasets. 
Consequently, the corresponding SRS derived from these time series inherently reflect these constraints. 
This standardization guarantees that all evaluated SRS representations are directly comparable, operate within a consistent temporal–spectral bandwidth, and remain fully congruent with the data distribution used to train the ML model.

\paragraph{Performance Metrics}
There are no established metrics in the shock or SRS reconstruction domain that comprehensively evaluate model accuracy, robustness, and fidelity across both global and local spectral domains. 
To address this, we developed two complementary metrics designed to assess reconstruction performance from multiple perspectives:

\begin{itemize}
    \item \textbf{Root-Mean-Square Logarithmic Error (RMSLE):}  
    A scalar, distribution-level measure of overall spectral agreement between the reconstructed and target SRS. 
    It captures both magnitude and proportional deviations across the frequency range, serving as an absolute indicator of global performance. 
    
    In addition, we report a per-sample (one-to-one) RMSLE accuracy metric that compares the RMSLE of the ML-generated SRS to that of the SDS baseline for each target spectrum. 
    When evaluated over a large ensemble of test samples, this comparison yields an empirical estimate of the probability of relative performance, quantifying the likelihood that the ML model achieves lower RMSLE than the baseline on a randomly drawn test case.
    
    \item \textbf{Per-Frequency dB Error:}  
    A localized frequency-domain measure that assesses deviations at each spectral component on a logarithmic scale ($20\log_{10}$).  
    This enables both visual and statistical interpretation of the magnitude and spread of reconstruction errors relative to practical engineering thresholds (e.g., $\pm3$~dB).
\end{itemize}

Together, these metrics provide a holistic characterization of model performance, capturing both global fidelity and local spectral accuracy.

\begin{figure}[H]
\centering
\includegraphics[page=3, trim=0cm 3cm 14cm 0cm, clip, width=0.5\linewidth]{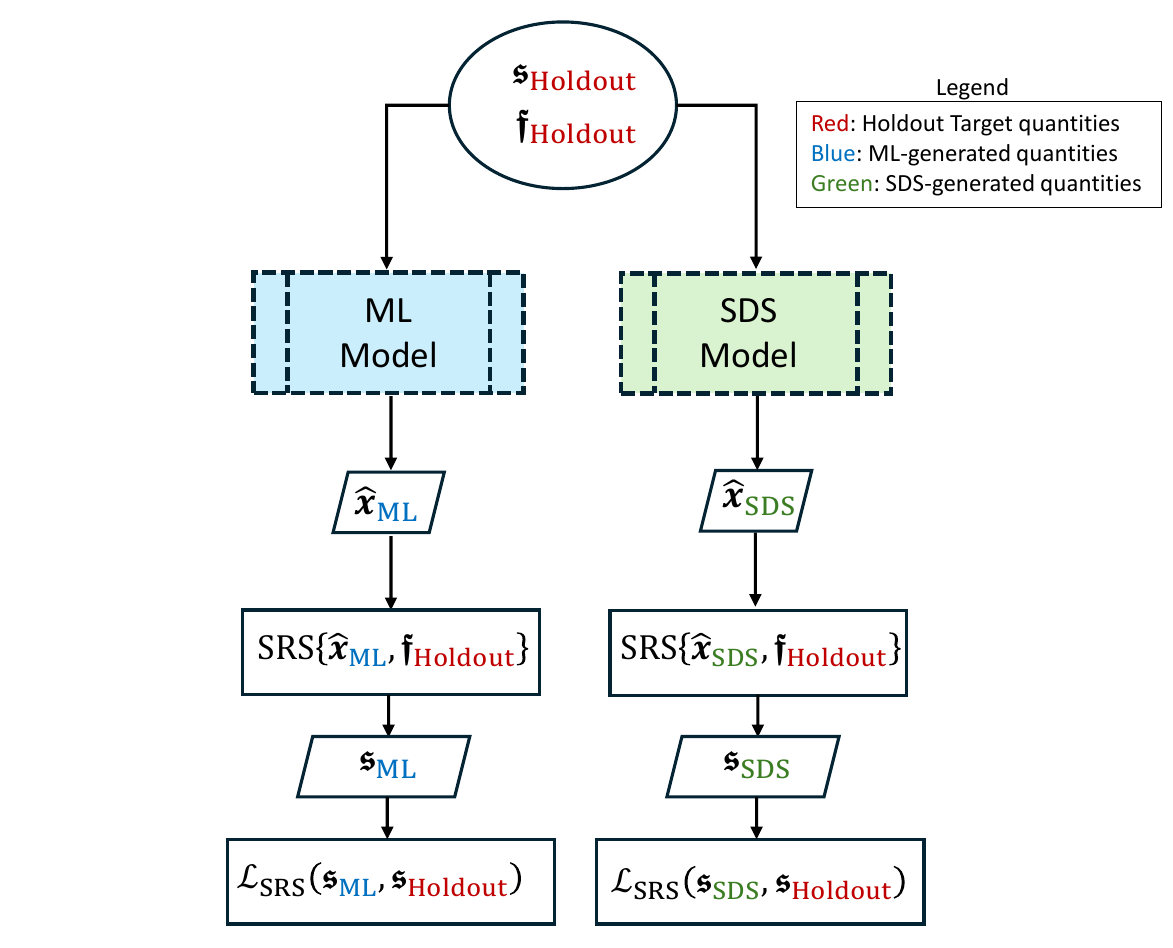}
\caption{Evaluation workflow for the machine learning (ML) and classical sum-of-decayed-sinusoids (SDS) inverse SRS models on the hold-out datasets. 
Each model receives identical hold-out target spectra and evaluation frequencies $(\bm{\mathfrak{s}}_{\mathrm{Holdout}}, \bm{\mathfrak{f}}_{\mathrm{Holdout}})$ as inputs and generates a candidate acceleration time series $\hat{x}_{\mathrm{ML}}$ or $\hat{x}_{\mathrm{SDS}}$. 
The SRS of each reconstructed signal is then computed and compared to the target spectrum using the SRS loss $\mathcal{L}_{\mathrm{SRS}}$ providing a RSMLE error for each methodology. 
The resulting RMSLE errors across entire held-out data sets are visualized in Figure~\ref{fig:Holdout_RMSLE_ABCD}, quantify reconstruction accuracy and generalization performance across independent hold-out datasets.}
\label{fig:holdout_srs_framework}
\end{figure}

\begin{figure}[H]
\centering
\includegraphics[width=0.9\textwidth]{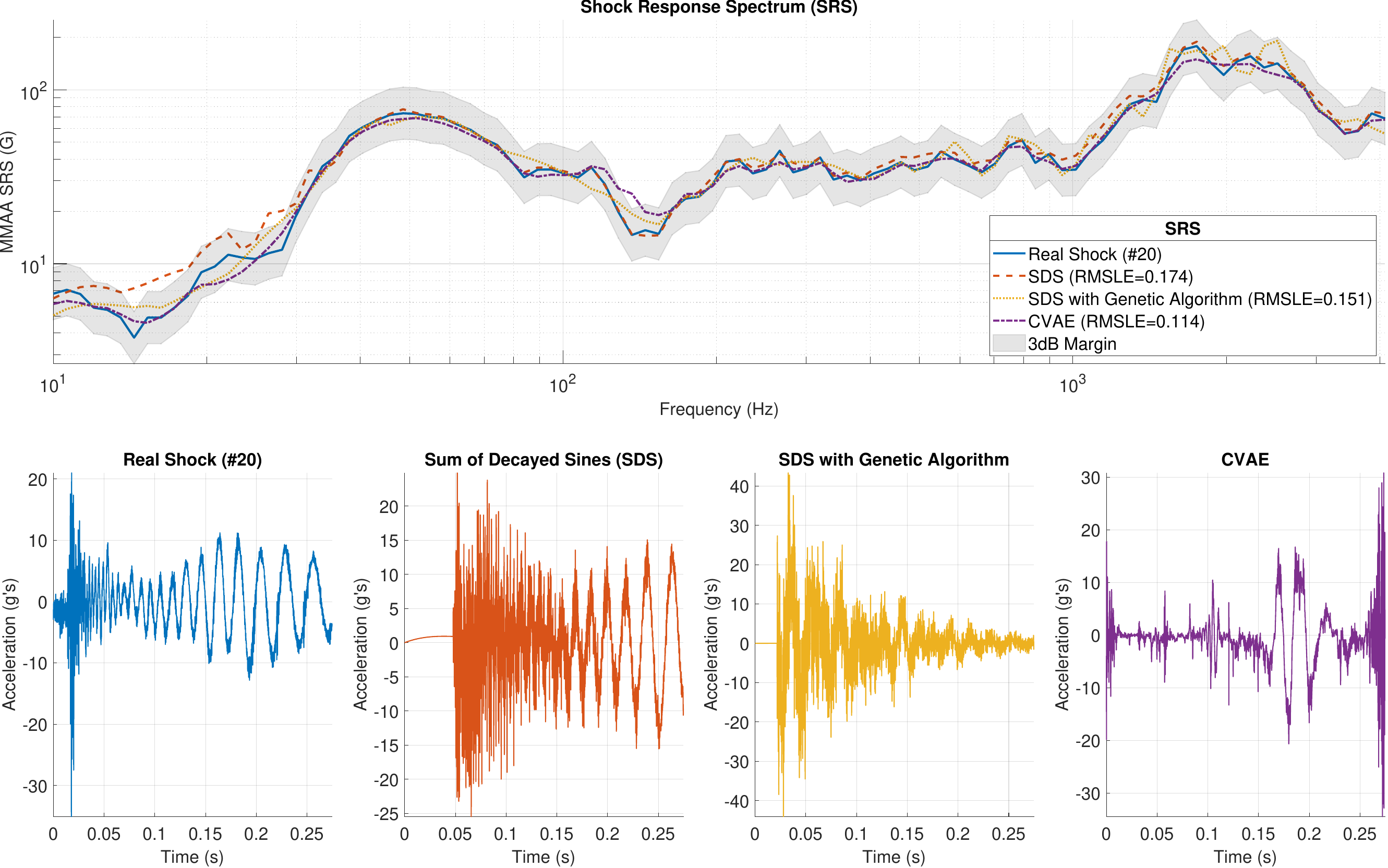}
\caption{Example comparison from the auxiliary held-out dataset $\mathcal{D}_{\text{test}}^{\text{A}}$. 
Top: shock response spectrum (SRS) of a real shock (black) with reconstructions obtained using classical SDS (orange), 
SDS with a genetic algorithm (yellow), and the proposed CVAE (purple). 
Bottom: corresponding time-domain signals. 
All methods reproduce the broad spectral shape; CVAE reduces RMSLE relative to classical SDS, 
while SDS+GA achieves the lowest overall error at substantially greater runtime.}
\label{fig:sds_vs_ML}
\end{figure}

\begin{figure}[H] 
    \begin{subfigure}{0.49\textwidth}
        \centering
        \includegraphics[width=1.0\textwidth]{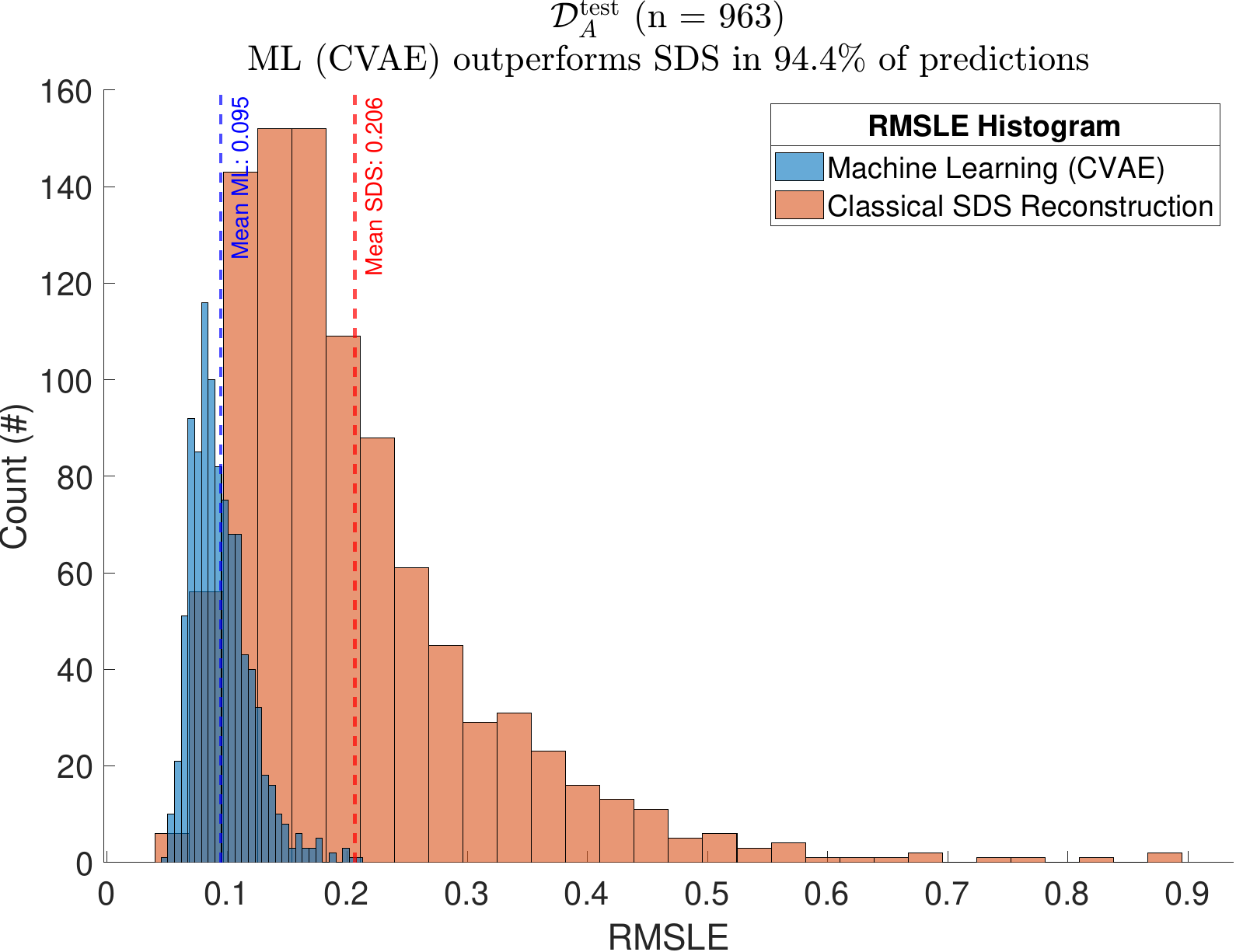}
    \end{subfigure}
        \vspace{1em}
    \begin{subfigure}{0.49\textwidth}
        \centering
        \includegraphics[width=1.0\textwidth]{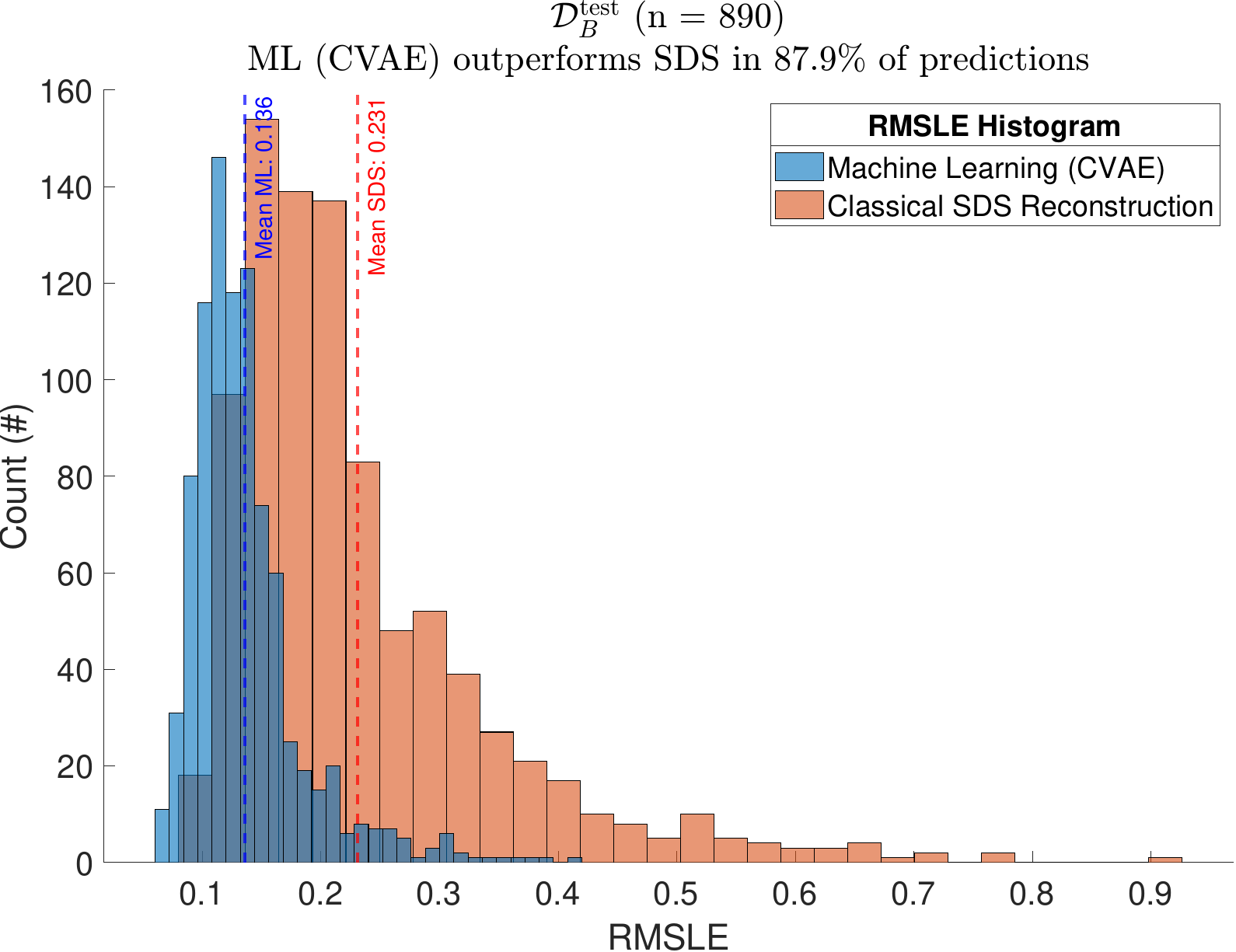} 
    \end{subfigure}
    \begin{subfigure}{0.49\textwidth}
        \centering
        \includegraphics[width=1.0\textwidth]{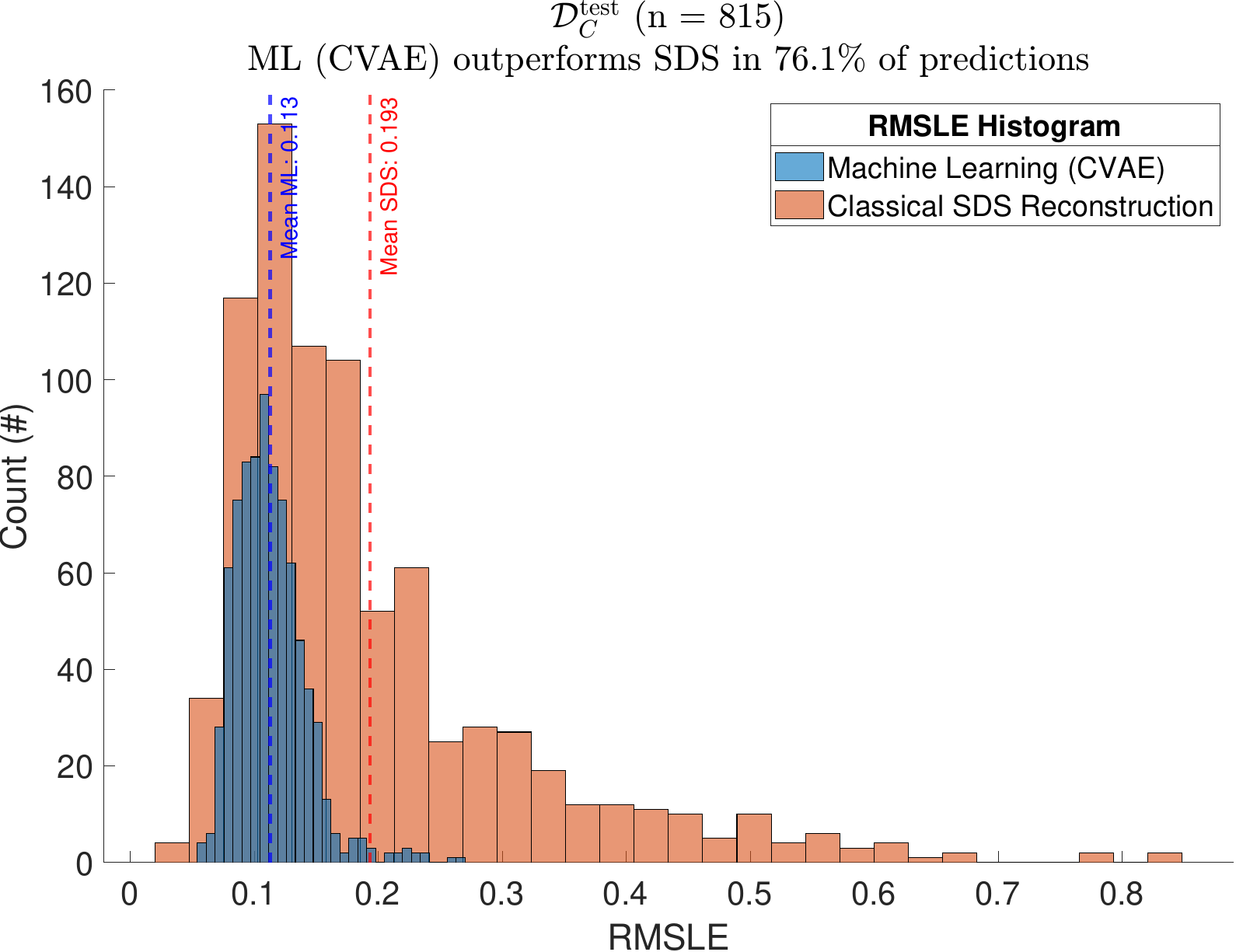} 
    \end{subfigure}
    \begin{subfigure}{0.49\textwidth}
        \centering
        \includegraphics[width=1.0\textwidth]{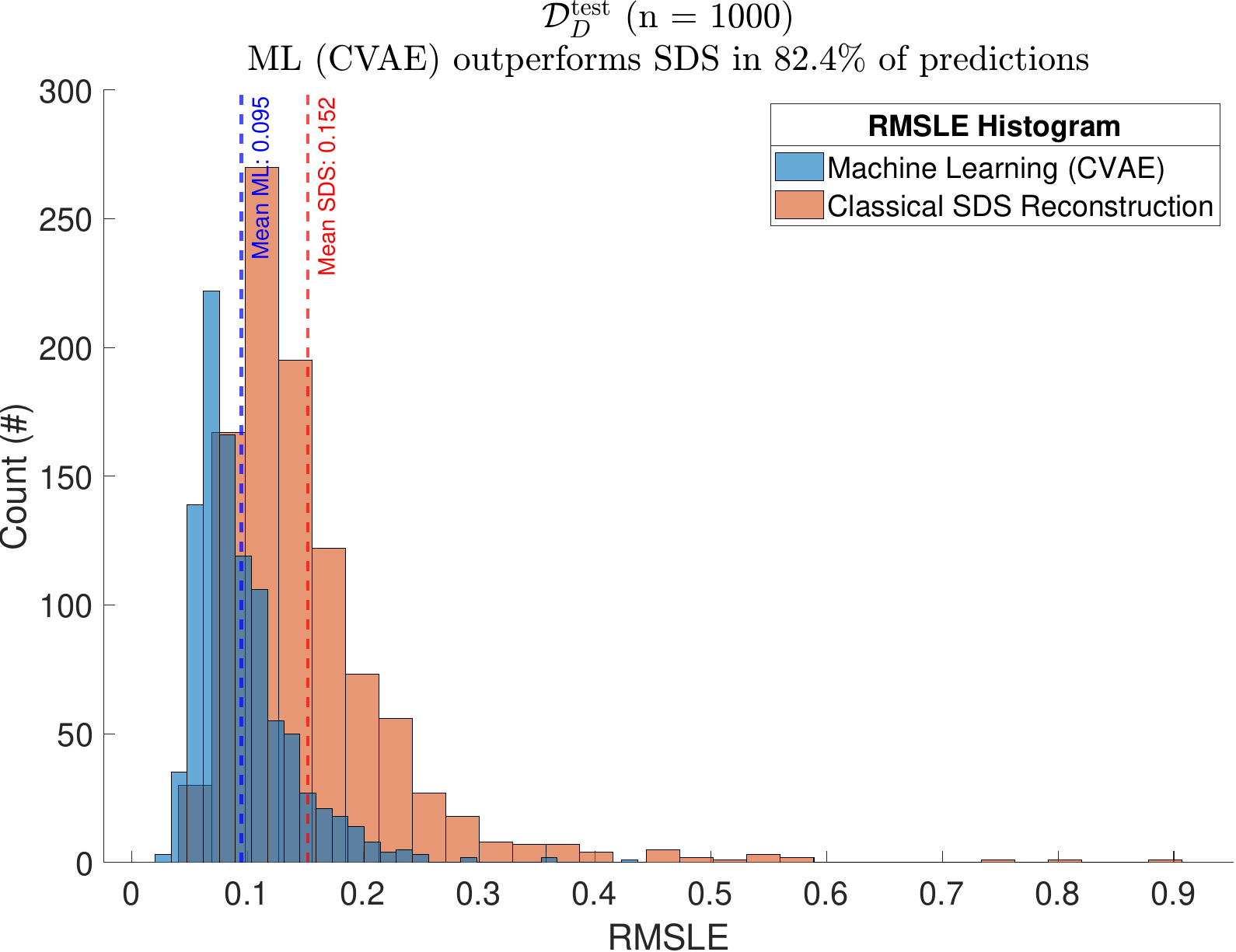} 
    \end{subfigure}
    \caption{Distribution of reconstruction errors (RMSLE) comparing CVAE-generated shocks (blue) and classical SDS reconstructions (red) across four independent held-out test datasets. 
Top left: $\mathcal{D}_{\text{test}}^{\text{A}}$ ($n=976$), where the CVAE outperforms SDS in 94.4\% of predictions. 
Top right: $\mathcal{D}_{\text{test}}^{\text{B}}$ ($n=890$), where the CVAE achieves lower error in 87.9\% of cases. 
Bottom left: $\mathcal{D}_{\text{test}}^{\text{C}}$ ($n=815$), where the CVAE exceeds SDS performance in 76.1\% of predictions. 
Bottom right: $\mathcal{D}_{\text{test}}^{\text{D}}$ ($n=1000$), where the CVAE outperforms SDS in 82.4\% of cases. 
These results demonstrate that while the CVAE generally improves reconstruction accuracy, its performance may degrade in underrepresented or highly complex shock regimes.}
    \label{fig:Holdout_RMSLE_ABCD}
\end{figure}

\paragraph{Root-Mean-Square Logarithmic Error (RMSLE)}
Figure~\ref{fig:sds_vs_ML} also provides the reconstructed time series obtained using the SDS method $(\hat{x}_{\mathrm{SDS}})$ and the CVAE model $(\hat{x}_{\mathrm{ML}})$, along with their corresponding SRS responses. 
The reported RMSLE values quantify the deviation between the target SRS and those produced by the SDS and ML reconstructions. 
The workflow used to generate the overall distribution of these results is summarized in Figure~\ref{fig:holdout_srs_framework}, while Figure~\ref{fig:Holdout_RMSLE_ABCD} shows the RMSLE distributions across the four independent held-out datasets with sample sizes $n_{\text{A}}=976$, $n_{\text{B}}=890$, $n_{\text{C}}=815$, and $n_{\text{D}}=1000$. 

To evaluate model behavior at the individual-case level, we perform a one-to-one comparison for each target spectrum $j$ in a hold-out set using the indicator, $\mathbf{1} \!=\! [\mathrm{RMSLE}_{\mathrm{ML}}^{(j)} < \mathrm{RMSLE}_{\mathrm{SDS}}^{(j)}]$. 
The per-sample accuracy is then the fraction of cases where the ML reconstruction attains lower error than SDS. 
Across the four held-out datasets, the CVAE outperforms SDS on the majority of samples, achieving win rates of 94.4\%, 87.9\%, 76.1\%, and 82.4\% for $\mathcal{D}_{\text{test}}^{A}$ through $\mathcal{D}_{\text{test}}^{D}$, respectively. 
This per-sample perspective complements the distribution-level RMSLE analysis and the localized per-frequency dB Error view: it is less influenced by large-error outliers that may inflate the mean and directly reflects how often the ML method yields a more accurate reconstruction for a given target SRS.

Table~\ref{tab:rmsle_summary} summarizes the descriptive statistics of RMSLE for both the CVAE and classical SDS methods across all datasets. 
The CVAE exhibits lower (better) median RMSLE values across \textit{all} held-out datasets. The CVAE exhibits favorable minimum, maximum, standard deviation, and quantile statistics across the four held-out datasets indicating consistently smaller and tighter error distributions. 

Taken together, these results demonstrate that the CVAE provides more accurate and stable reconstructions across most test conditions, outperforming SDS in overall error magnitude and consistency.

\begin{table}[h]
\centering
\renewcommand{\arraystretch}{1.2}
\setlength{\tabcolsep}{6pt}
\caption{Summary statistics of RMSLE for machine learning (ML) and classical SDS reconstructions across four independent held-out datasets.
Reported metrics include the mean, median, standard deviation, minimum, maximum, and selected quantiles.
Bold values indicate the lower (better) \textbf{Mean}, \textbf{Median}, or \textbf{Standard Deviation} within each dataset.}
\label{tab:rmsle_summary}
\begin{tabular}{ccccccccc}
\toprule
\textbf{Hold-out} & \textbf{Method} & \textbf{Mean} & \textbf{Median} & \textbf{Std. Dev.} & \textbf{Min} & \textbf{Max} & \textbf{0.025 Quant.} & \textbf{0.975 Quant.}  \\
\midrule
A & SDS & 0.206 & 0.176 & 0.11 & 0.051 & 0.894 & 0.082 & 0.489 \\
A & ML  & \textbf{0.095} & \textbf{0.090} & \textbf{0.024} & 0.049 & 0.211 & 0.060 & 0.156 \\
\arrayrulecolor{gray!30}\midrule
\arrayrulecolor{black}
B & SDS & 0.231 & 0.200 & 0.112 & 0.088 & 0.925 & 0.110 & 0.546 \\
B & ML  & \textbf{0.136} & \textbf{0.125} & \textbf{0.047} & 0.063 & 0.418 & 0.079 & 0.266 \\
\arrayrulecolor{gray!30}\midrule
\arrayrulecolor{black}
C & SDS & 0.193 & 0.154 & 0.123 & 0.038 & 0.846 & 0.067 & 0.544 \\
C & ML  & \textbf{0.113} & \textbf{0.109} & \textbf{0.029} & 0.059 & 0.269 & 0.072 & 0.186 \\
\arrayrulecolor{gray!30}\midrule
\arrayrulecolor{black}
D & SDS & 0.152 & 0.131 & 0.081 & 0.046 & 0.907 & 0.068 & 0.369 \\
D & ML  & \textbf{0.095} & \textbf{0.083} & \textbf{0.042} & 0.028 & 0.437 & 0.044 & 0.201 \\
\bottomrule
\end{tabular}
\end{table}

\paragraph{Per-Frequency dB Error}
The per-frequency dB error, evaluated in decibel space via a $20\log_{10}(\cdot)$ magnitude transform, provides a localized measure of reconstruction fidelity across the SRS spectrum.
Rather than aggregating across frequencies, this metric evaluates the deviation at each frequency point, thereby capturing spectral regions where reconstruction errors may concentrate. 
For most shock and vibration applications, deviations within $\pm3$~dB are considered acceptable.

\begin{figure}[H] 
    \begin{subfigure}{0.49\textwidth}
        \centering
        \includegraphics[width=1.0\textwidth]{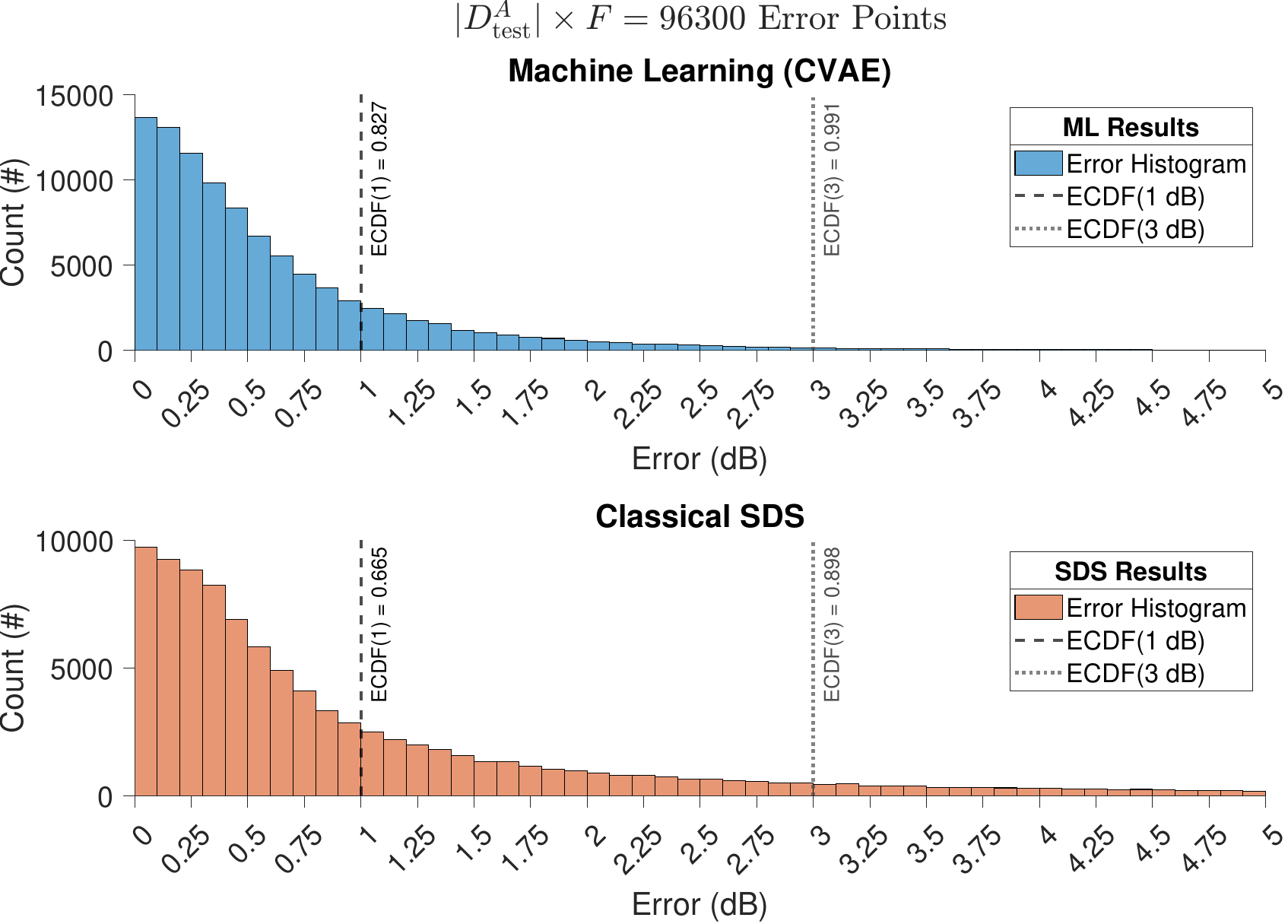}
    \end{subfigure}
    \vspace{1em}
    \begin{subfigure}{0.49\textwidth}
        \centering
        \includegraphics[width=1.0\textwidth]{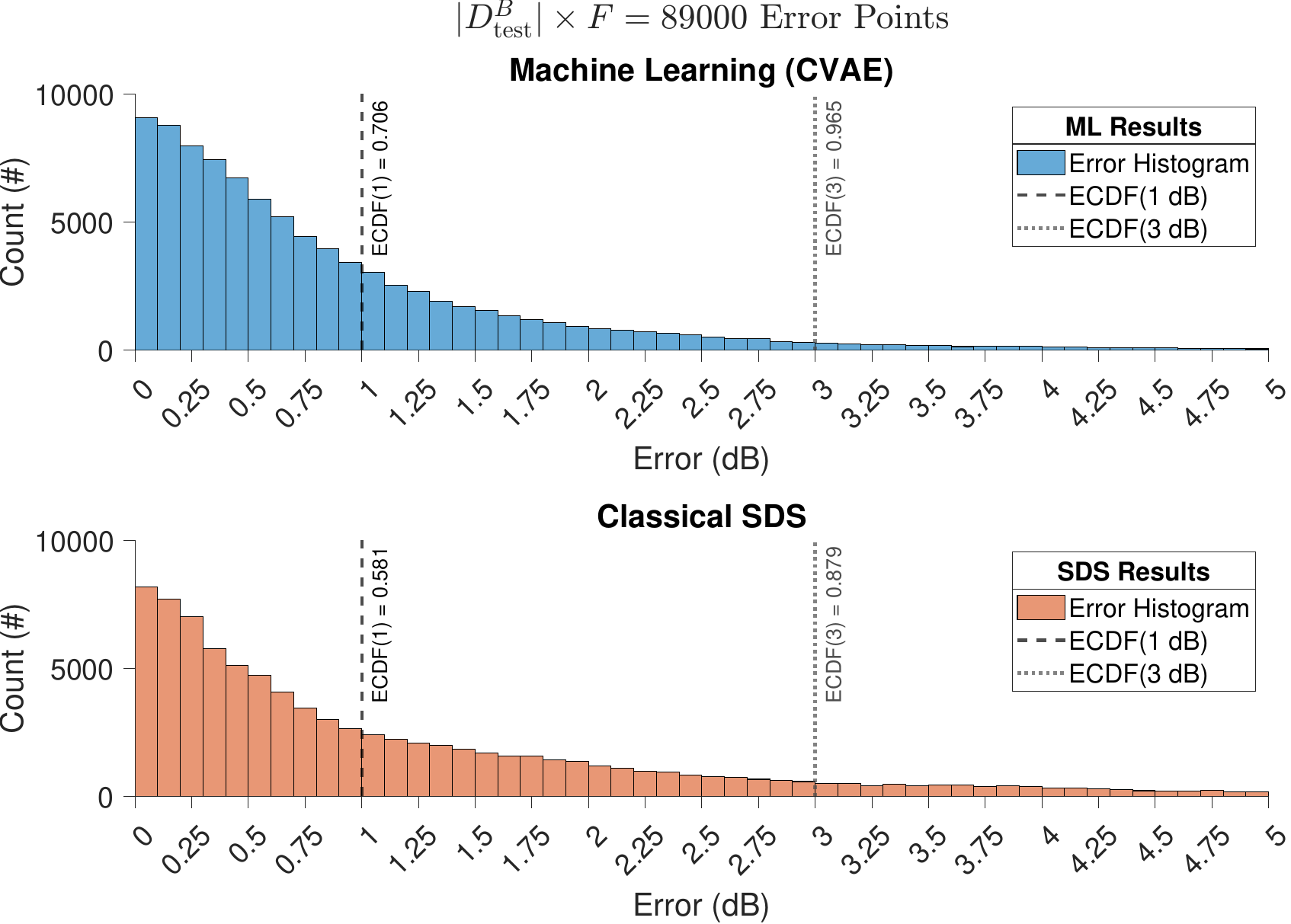} 
    \end{subfigure}
    \begin{subfigure}{0.49\textwidth}
        \centering
        \includegraphics[width=1.0\textwidth]{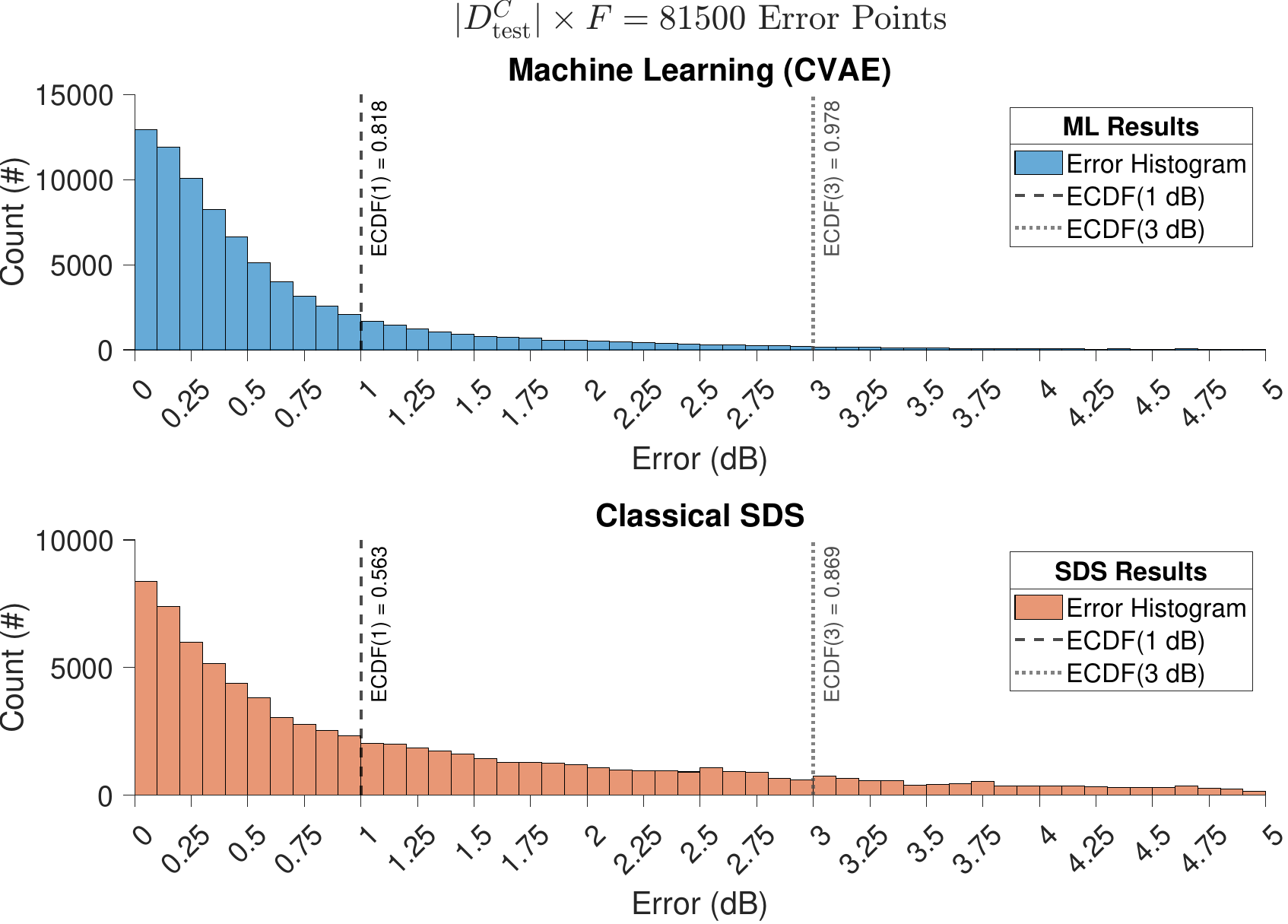} 
    \end{subfigure}
    \begin{subfigure}{0.49\textwidth}
        \centering
        \includegraphics[width=1.0\textwidth]{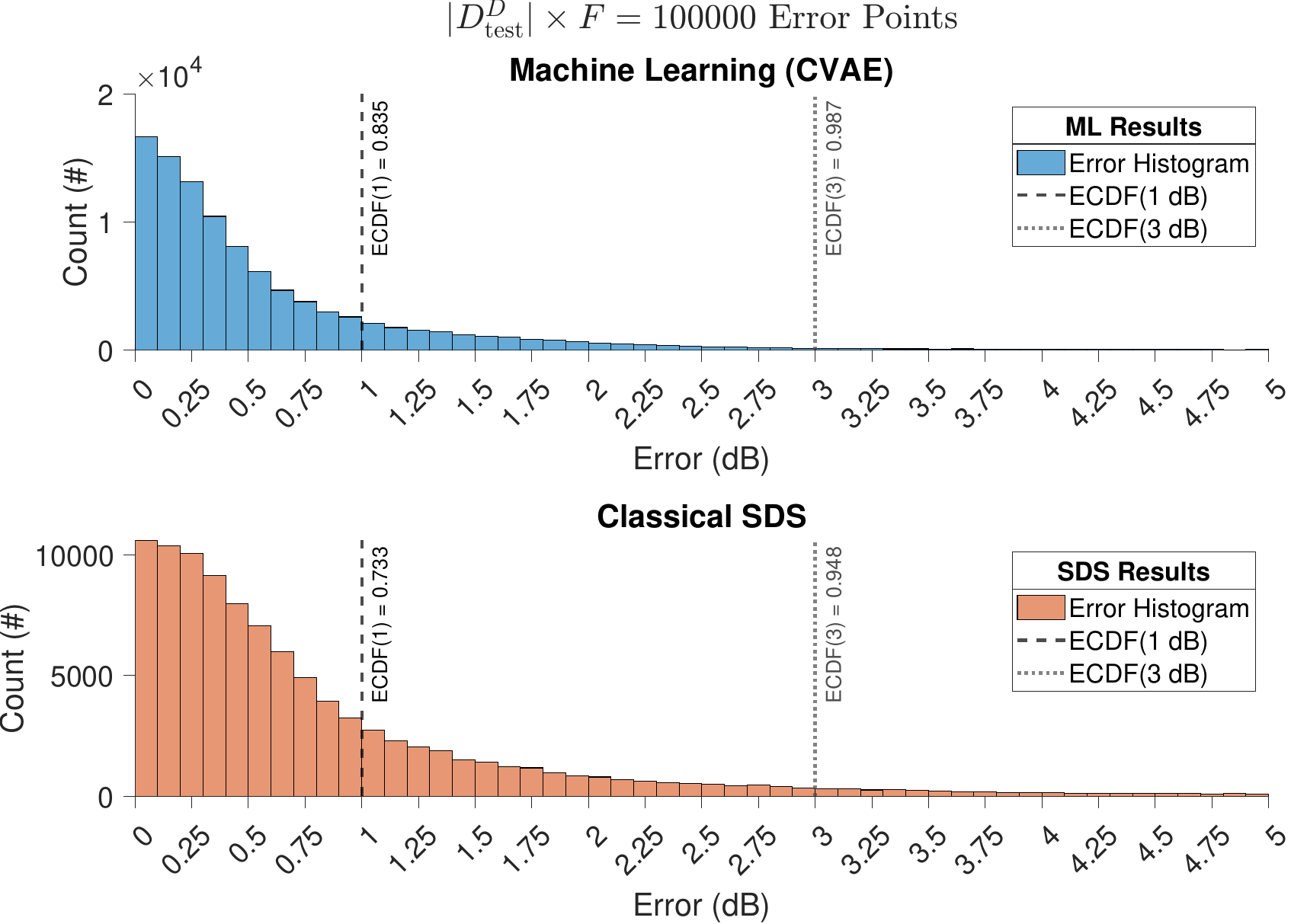} 
    \end{subfigure}
    \caption{
    Comparison of machine learning (CVAE) and classical SDS reconstruction errors across the four held-out test datasets ($\mathcal{D}_{\text{test}}^{A}$ through $\mathcal{D}_{\text{test}}^{D}$). 
    Each subplot shows the histogram of SRS magnitude errors (in dB) along with the empirical cumulative distribution functions (ECDF) for 1~dB and 3~dB thresholds. 
    The number of evaluated SRS points ($|\mathcal{D}_{\text{test}}| \times F$) is indicated above each panel. 
    Overall, the CVAE achieves lower error distributions and higher fractions of points within 1~dB and 3~dB tolerance compared to the SDS method, across all holdout sets.
    }
    \label{fig:Holdout_dbEror_ABCD}
\end{figure}

Figure~\ref{fig:Holdout_dbEror_ABCD} compares the SRS reconstruction errors produced by the CVAE model and the SDS method across the four held-out datasets ($\mathcal{D}_{\text{test}}^{A}$ through $\mathcal{D}_{\text{test}}^{D}$). 
Each subplot shows histograms of per-frequency SRS magnitude errors (in dB) alongside the empirical cumulative distribution functions (ECDF) for 1~dB and 3~dB thresholds.
These distributions are computed over all evaluation natural frequencies, $\boldsymbol{\mathfrak{f}}$, for every sample in the corresponding held-out dataset. This can be represented as $|\mathcal{D}_j| \times \ |\bm{\mathfrak{f}}| \quad j \in \{A, B, C, D\}$, yielding between $8\times10^{4}$ and $1\times10^{5}$ total error points per hold-out set.

Across all of the held-out datasets, the CVAE demonstrates consistently lower reconstruction errors relative to SDS, with the histogram indicating a higher proportion of low-error points within both the 1~dB and 3~dB ranges. 

Overall, results across all metrics favor the CVAE: it achieves competitive or superior accuracy in most cases while offering orders-of-magnitude faster inference time than optimization-based reconstruction methods. 
These findings demonstrate that the learned inverse mapping generalizes robustly to unseen shocks and maintains strong performance across diverse spectral and temporal conditions.

\section{Discussion}\label{sec:Discussion}

Figure~\ref{fig:Holdout_dbEror_ABCD} shows that our CVAE was able to generate shocks whose SRS matched the targets with high fidelity, typically within $\pm 3$~dB.
Additionally a higher percentage of the dB error for our CVAE  was within the 1dB tolerance in the ECDF, indicating that more of the CVAE's error distribution was within 1dB of the SRS as opposed to SDS. 
Additionally, Figure~\ref{fig:Holdout_RMSLE_ABCD} shows that our RMSLE was consistently lower than the SDS RMSLE across the 4 holdouts, often by sizable margins.
These two metrics, RMSLE and per-frequency dB error—provide complementary perspectives on reconstruction performance, offering a practical framework for assessing both accuracy and consistency. 

Figure~\ref{fig:sds_vs_ML} offers a per-sample view of these trends using a representative comparison from $\mathcal{D}_{\text{test}}^{\text{A}}$, showing results for the CVAE, SDS, and SDS augmented with a genetic algorithm (SDS+GA).
The RMSLE values between each method’s reconstructed SRS and the target SRS were $0.114$ for the CVAE, $0.174$ for SDS, and $0.151$ for SDS+GA, demonstrating that the ML approach achieves the highest accuracy while being orders of magnitude faster at inference.

Overall, the CVAE generalized well and consistently outperformed the classical sum-of-decayed-sinusoids (SDS) baseline across both of the evaluation metrics and hold-out sets. 

Currently, there are no established benchmarks for systematic comparison of SRS reconstruction methods, underscoring the need for standardized evaluation practices. 
The curated hold-out datasets released with this study are intended to help fill this gap by providing a standardized, diverse, and reproducible testbed for evaluating and comparing future SRS inversion and shock synthesis approaches.

\subsection{Encoding and Computational Efficiency}

The frequency reweighting and logarithmic transformation in Eq.~\eqref{EQ:Freq_Encoding} represent only a first-pass approach to embedding frequency information. 
While this strategy performed adequately for the present study, more sophisticated encoding schemes—such as Feature-wise Linear Modulation (FiLM) or learned frequency embeddings—could provide richer and more flexible representations, potentially leading to further performance gains \cite{perez2018film}.

A key advantage of the CVAE is its exceptional computational efficiency. 
Reconstructing a single shock time series required only about $0.3$~ms---compared with $5$--$8$~s for SDS and $20$--$30$~min for SDS+GA. 
The CVAE delivered superior accuracy while being four to six orders of magnitude faster at inference. 
This dramatic speedup makes the approach well suited for large-scale simulations and real-time applications. 
Furthermore, such low inference cost enables ensemble learning strategies~\cite{sagi2018ensemble}, where multiple lightweight models can generate diverse candidate time series consistent with a given target SRS. 
With continued architectural refinement, both accuracy and diversity could be further improved without compromising the remarkable efficiency demonstrated here.




\section{Conclusion}\label{sec:Conclusion}

This work demonstrated that a conditional variational autoencoder (CVAE) can serve as an effective framework for shock response spectrum (SRS) inversion. 
The trained model successfully generated shock acceleration time series with high fidelity to their target SRS, eliminating the need for explicit basis functions required in classical approaches. 
By directly learning the underlying data distribution, the CVAE produced realistic signals that generalized well across a diverse range of SRS conditions. 
Model performance was evaluated using two complementary metrics introduced in this study: RMSLE, with per-sample accuracy, and per-frequency dB error. 
Together, these metrics demonstrated strong reconstruction accuracy and robustness relative to traditional methods. 
A major advantage of the proposed approach is its exceptional computational efficiency, as inference requires only a few milliseconds per realization, representing orders-of-magnitude speedups compared to optimization-based techniques such as SDS and SDS+GA.

Despite these strengths, several challenges remain. 
The generated time series exhibited limited diversity, suggesting that the latent space is underutilized. 
This issue could potentially be mitigated through richer latent priors, hierarchical encoders, or more expressive model architectures. 
Additionally, the current CVAE implementation is constrained to shock durations of $\approx 0.2747$~s, which limits applicability when multiple shocks occur in sequence or when both low- and high-frequency content must be captured simultaneously. 
Such cases demand longer time windows and high sampling rates, which may become feasible with advances in computational hardware. 
It is reasonable to expect that extending the sequence length from 9{,}000 to 90{,}000 samples (approximately 2.7~s) would allow the model to represent nearly all practical shock scenarios.

In addition to proposing the CVAE-based inversion framework, this work introduced a large-scale synthetic shock data generation pipeline capable of producing hundreds of thousands of physically consistent training samples with controlled spectral diversity. 
We further established a set of standardized hold-out datasets that enable consistent and reproducible evaluation of model generalization—helping address the current absence of benchmarks in SRS reconstruction research. 
Together, these contributions provide a foundation for continued progress in data-driven shock synthesis and model-based inversion. 

Overall, this study demonstrates that machine learning, and CVAEs in particular, offer a powerful and computationally efficient approach to the SRS inverse problem. 
With continued refinement, such models have strong potential to complement or replace traditional methods in large-scale simulation, qualification testing, and real-time control applications.

\bibliographystyle{IEEEtran}
\bibliography{GenerativeTimeSeries}

\clearpage
\appendix

\section*{Notation}
\addcontentsline{toc}{section}{Notation} 

\subsection*{Mechanical model and SRS}

\begin{table}[htbp]
\centering
\renewcommand{\arraystretch}{1.3}
\begin{tabularx}{\textwidth}{M X}
\toprule
\textbf{Symbol} & \textbf{Definition} \\
\midrule
m & Mass of the SDOF oscillator. \\
k_i & Spring stiffness of the $i$-th SDOF oscillator. \\
c_i & Viscous damping of the $i$-th SDOF oscillator. \\
x(t) & Base displacement (input excitation). \\
y(t) & Absolute displacement of the mass. \\
z(t)=y(t)-x(t) & Relative displacement of the mass w.r.t.\ the base. \\
\dot{x}(t),\,\ddot{x}(t) & Base velocity and base acceleration (shock). \\
\dot{y}(t),\,\ddot{y}(t) & Absolute velocity and acceleration of the mass. \\
\omega_i=\sqrt{k_i/m} &  Undamped natural angular frequency (rad/s) of the $i$-th SDOF oscillator. \\
\mathfrak{f}_i=\omega_i/(2\pi) &  Natural frequency (Hz) of the $i$-th SDOF oscillator. \\
\zeta= 0.03 = c_i/(2m\omega_i) & Constant damping ratio across all SDOF oscillators. \\
\omega_d=\omega_i\sqrt{1-\zeta^2} & Damped natural angular frequency (rad/s) of the $i$-th SDOF oscillator. \\
\bottomrule
\end{tabularx}
\caption{Notation: Mechanical model of the base-excited SDOF oscillator.}
\label{tab:notation_mech}
\end{table}

\subsection*{Continuous and Discrete Time Series and SRS}
\begin{table}[H]
\centering
\renewcommand{\arraystretch}{1.2}
\begin{tabularx}{\textwidth}{M X}
\toprule
\textbf{Symbol} & \textbf{Definition} \\
\midrule
n & Discrete-time index. \\
N\!=\!9000 & Number of discrete-time indices (total number of samples). \\

\ddot{x}(t) & Continuous-time acceleration signal (input shock). \\

\ddot{x}[n] & Discrete-time acceleration signal obtained by sampling $\ddot{x}(t)$. \\

\ddot{x}^{(j)}(t) & $j$-th continuous-time acceleration realization. \\

\ddot{x}^{(j)}[n] & $j$-th discrete-time acceleration realization. \\

\ddot{\boldsymbol{x}}(t) 
& Collection (tensor) of continuous-time acceleration realizations. \\

\ddot{\boldsymbol{x}}[n] 
& Collection (tensor) of discrete-time acceleration realizations,
$\in \mathbb{R}^{J \times N}$. \\

\hat{x}_{\mathrm{SDS}}[n] & Reconstructed discrete-time series using the sum of decayed sinusoids (SDS). \\
\hat{x}_{\mathrm{ML}}[n] & Reconstructed discrete-time series using the ML model (CVAE). \\

f_s\!=\!32768 & Sampling frequency (Hz). \\
T_s\!=\!1/f_s & Sampling period (s). \\
\mathfrak{f}_{\min}\!=\!10 & Minimum evaluation natural frequency used in the SRS calculation (Hz). \\
\mathfrak{f}_{\max}\!=\!4096 & Maximum evaluation natural frequency used in the SRS calculation (Hz). \\
\bm{\mathfrak{f}} = [\mathfrak{f}_{\min}, \ldots, \mathfrak{f}_{\max}] & Vector of evaluation natural frequencies used in the SRS calculation (Hz). \\
\mathfrak{f}_i & $i$-th evaluation natural frequency (Hz), element of $\bm{\mathfrak{f}}$ defined in Table~\ref{tab:notation_mech}. \\
F\!=\!100 & Cardinality of evaluation natural frequencies used in SRS operator, $|\bm{\mathfrak{f}}|$. \\
L\!=\!1640 & Zero-padding length used in SRS computation (samples). \\
p\!=\!3 & Padding scale factor used in SRS computation, where padding is reduced from $L$ to $L/p$. \\
\mathrm{SRS}\!\left\{\ddot{x},\bm{\mathfrak{f}};\zeta \right\} & SRS operator evaluated at $\bm{\mathfrak{f}}$, with damping ratio $\zeta$ assumed constant when omitted. \\
\bm{\mathfrak{s}}_{x} & Output of the SRS operator, i.e., SRS magnitudes $\bm{\mathfrak{s}}_{x} \in \mathbb{R}^{F}$. \\
\boldsymbol{\mathfrak{S}} & Matrix of stacked SRS vectors across realizations, $\boldsymbol{\mathfrak{S}} \in \mathbb{R}^{J \times F}$. \\
\bottomrule
\end{tabularx}

\caption{Notation used for continuous and discrete-time signal representation and shock response spectrum (SRS) computation. 
A superscript $(j)$ denotes a particular instance, while $n$ indexes discrete time. 
The conventional double-dot notation for acceleration ($\ddot{x}$) is suppressed for clarity in reconstructed signals. Cardinality of a set is denoted as $|\bullet|$ . }
\label{tab:notation_srs}
\end{table}

\subsection*{Basis functions and synthetic generation}
\begin{table}[H]
\centering
\renewcommand{\arraystretch}{1.2}
\begin{tabularx}{\textwidth}{M X}
\toprule
\textbf{Symbol} & \textbf{Definition} \\
\midrule
M & Number of sinusoidal basis functions in the SDS model. \\
A_i & Amplitude of the $i$-th sinusoidal basis function (optimized in SDS reconstruction). \\
f_i & Frequency of the $i$-th sinusoidal basis function (optimized in SDS reconstruction). \\
\lambda_i & Exponential decay constant of the $i$-th sinusoidal basis function (optimized in SDS reconstruction). \\
\phi_i & Phase of the $i$-th sinusoidal basis function (optimized in SDS reconstruction). \\
\mathcal{L}^{\text{SDS}} _{\text{SRS}} & RMSLE loss comparing target and reconstructed SDS SRS. \\
\hat{\boldsymbol{\theta}}_{\mathrm{SDS}} & Estimated SDS parameter vector obtained by 
$\underset{\boldsymbol{\theta}\in\boldsymbol{\Theta}}{\arg\min}\!\left( \mathcal{L}^{\text{SDS}} _{\text{SRS}} \right)$. \\
\bottomrule
\end{tabularx}
\caption{Notation: Parameters and loss function for the sum of decayed sinusoids (SDS) model. 
Parameters $A_i, f_i, \lambda_i,$ and $\phi_i$ represent the sinusoidal basis functions 
and are estimated through optimization in SDS reconstruction. 
Here the double-dot notation for acceleration ($\ddot{x}$) is suppressed for simplicity.}
\label{tab:notation_sds}
\end{table}

\clearpage
\subsection*{Synthetic shock generation: parameters and distributions}

\renewcommand{\arraystretch}{1.22}
\begin{table}[H]
\centering
\small
\begin{tabularx}{\textwidth}{M>{\raggedright\arraybackslash}X M>{\raggedright\arraybackslash}p{0.18\textwidth}}
\toprule
\textbf{Parameter} & \textbf{Definition} & \textbf{Distribution} & \textbf{Distribution Name} \\
\midrule
\ddot{x}_{\text{synth}}[n] = \epsilon + \sum_{i=1}^{\mathcal{B}} \psi_i[n] & One discrete-time synthetic shock signal of length $N$, generated by summing $\mathcal{B}$ basis functions and noise. & -- & -- \\
\mathcal{B} & Number of basis functions (sinusoid and/or wavelet) used in synthetic generation & \mathcal{U}\{1,10\} & Discrete uniform \\
\psi_i[n] & $i$-th basis function, which may be either a sinusoidal atom or a wavelet atom & -- & -- \\
\midrule
\textbf{Timing / placement} & & & \\
\xi_i & Fractional offset of the $i$-th basis function (start index $\lceil \xi_i N \rceil$) & U(0,\,0.75) & Continuous uniform \\
\gamma & Adoption rate for offsets ($\xi_{i+1}\!\leftarrow\!\xi_i$ with prob.\ $\gamma$) & \mathrm{Bernoulli}(p=0.5) & Bernoulli \\
\midrule
\textbf{Basis-function parameters (common)} & & & \\
A_i & Amplitude of the $i$-th basis function & U(0.25,\,10) & Continuous uniform \\
\phi_i & Phase of the $i$-th basis function & U(0,\,2\pi) & Continuous uniform \\
f_i & Basis frequency (Hz) & U(10,\,4096) & Continuous uniform \\
\midrule
\textbf{Sinusoid-specific} & & & \\
\lambda_i & Exponential decay constant of the $i$-th sinusoid & U(0.004\pi f_i,\,0.2\pi f_i) & Continuous uniform \\
\midrule
\textbf{Wavelet-specific} & & & \\
\eta & Damping/shape factor for wavelet atom & U(0.01,\,10) & Continuous uniform \\
\midrule
\textbf{Noise} & & & \\
\epsilon & Additive background noise & \mathcal{N}(0,\,\sigma_{\epsilon}^2) & Normal \\
\sigma_{\epsilon}^2 & Noise variance (sampled once and fixed per shock) & U(0.005,\,0.05) & Continuous uniform \\
\bottomrule
\end{tabularx}
\caption{Notation: Parameters and sampling rules used to generate synthetic shock data. 
Each synthetic signal $\ddot{x}_{\text{synth}}[n]$ is formed by summing $\mathcal{B}$ basis functions $\psi_i[n]$ 
(sinusoidal or wavelet) and adding background noise $\epsilon$. 
The dataset of $d$ such signals is denoted $\mathcal{D}_{\text{synth}} \in \mathbb{R}^{d \times N}$. 
Parameters $A_i, f_i, \lambda_i,$ and $\phi_i$ are consistent with the SDS model (Table~\ref{tab:notation_sds}), 
but here they are \emph{sampled} from distributions rather than optimized.}
\label{tab:DataGeneratorParams}
\end{table}
\renewcommand{\arraystretch}{1.0}

\clearpage
\subsection*{Datasets}

\renewcommand{\arraystretch}{1.22}
\begin{table}[H]
\centering
\small
\begin{tabularx}{\textwidth}{M X c}
\toprule
\textbf{Symbol} & \textbf{Definition} & \textbf{Cardinality} \\
\midrule
\mathcal{D}_{\text{synth}} \in \mathbb{R}^{d_{\text{synth}} \times N} 
& Synthetic dataset of generated shocks, each of length $N=9000$. & $|\mathcal{D}_{\text{synth}}| = 400{,}000$ \\
\mathcal{D}_{\text{real}} \in \mathbb{R}^{d_{\text{real}} \times N} 
& Real shock dataset of measured shocks, each of length $N=9000$. & $|\mathcal{D}_{\text{real}}| = 99{,}682$ \\
\mathcal{D}_{\text{combined}} \in \mathbb{R}^{(d_{\text{synth}}+d_{\text{real}})\times N} 
& Combined dataset containing both synthetic and real shocks. & $|\mathcal{D}_{\text{combined}}| = 499{,}682$ \\
\mathcal{D}_{\text{train}} \in \mathbb{R}^{d_{\text{train}} \times N} 
& Training split of $\mathcal{D}_{\text{combined}}$. & $|\mathcal{D}_{\text{train}}| = 494{,}685$ \\
\mathcal{D}_{\text{test}} \in \mathbb{R}^{d_{\text{test}} \times N} 
& Test split of $\mathcal{D}_{\text{combined}}$. & $|\mathcal{D}_{\text{test}}| = 4{,}997$ \\
\mathcal{D}_{\text{test}}^{\text{A}} \in \mathbb{R}^{d_a \times N} 
& Auxiliary held-out dataset for evaluation (not part of train/test split). & $|\mathcal{D}_{\text{test}}^{\text{A}}| = 976$ \\
\mathcal{D}_{\text{test}}^{\text{B}} \in \mathbb{R}^{d_b \times N} 
& Auxiliary held-out dataset for evaluation (not part of train/test split). & $|\mathcal{D}_{\text{test}}^{\text{B}}| = 890$ \\
\mathcal{D}_{\text{test}}^{\text{C}} \in \mathbb{R}^{d_c \times N} 
& Auxiliary held-out dataset for evaluation (not part of train/test split). & $|\mathcal{D}_{\text{test}}^{\text{C}}| = 815$ \\
\mathcal{D}_{\text{test}}^{\text{D}} \in \mathbb{R}^{d_d \times N} 
& Auxiliary held-out dataset for evaluation (not part of train/test split). & $|\mathcal{D}_{\text{test}}^{\text{D}}| = 1000$ \\
\bottomrule
\end{tabularx}
\caption{Notation: Datasets used in this study. Synthetic dataset $\mathcal{D}_{\text{synth}}$ is generated by the procedure in Table~\ref{tab:DataGeneratorParams}, 
$\mathcal{D}_{\text{real}}$ consists of measured shocks, and $\mathcal{D}_{\text{combined}}$ denotes their concatenation. 
This dataset is partitioned into training $\mathcal{D}_{\text{train}}$ (99\%) and test $\mathcal{D}_{\text{test}}$ (1\%). 
Four additional auxiliary held-out datasets, $\mathcal{D}_{\text{test}}^{\text{A}}$-$\mathcal{D}_{\text{test}}^{\text{D}}$, are used for independent evaluation only. 
All data are sampled uniformly at $f_s=32768$ Hz. Here $|\mathcal{D}|$ denotes the cardinality (number of shock time series) in dataset $\mathcal{D}$.}
\label{tab:datasets}
\end{table}
\renewcommand{\arraystretch}{1.0}

\subsection*{Machine learning (VAE/CVAE)}
\begin{table}[h]
\centering
\renewcommand{\arraystretch}{1.2}
\begin{tabularx}{\textwidth}{M X}
\toprule
\multicolumn{1}{c}{\textbf{Symbol}} & \textbf{Definition} \\
\midrule
\mathbf{x} & Discrete-time acceleration, i.e., $\mathbf{x} = \ddot{x}[n]$. \\
\mathbf{z} & Latent variable. \\
p(\mathbf{z}) & Prior over latent variables (usually $\mathcal{N}(0,\mathbf{I})$). \\
p_\theta(\mathbf{x}\mid \mathbf{z}) & Likelihood / decoder in VAE. \\
q_\phi(\mathbf{z}\mid \mathbf{x}) & Approximate posterior / encoder in VAE. \\
\boldsymbol{\mu}_q,\,\boldsymbol{\sigma}_q^2 & Parameters of the Gaussian approximate posterior $q_\phi(\mathbf{z}\mid \mathbf{x})$. \\
\mathbf{s} & Conditioning variable (SRS encoding at chosen frequencies). \\
q_\phi(\mathbf{z}\mid \mathbf{x},\mathbf{s}) & Encoder in CVAE (approximate posterior given data and context). \\
p_\theta(\mathbf{x}\mid \mathbf{z},\mathbf{s}) & Decoder in CVAE (conditional likelihood given latent and context). \\
p(\mathbf{z}\mid \mathbf{s}) & Conditional prior (typically $\mathcal{N}(0,\mathbf{I})$ in practice). \\
B & Batch size used in training. \\
W_{\mathfrak{f}_i}[n] & Frequency-dependent Gaussian weighting applied around the peak response (shape loss). \\
\text{SRS}_{\text{shape}}(\mathbf{x},\hat{\mathbf{x}}) & Weighted MSE between aligned SDOF responses (per batch and frequency); see Eq.~\eqref{eq:srs_shape}. \\
\mathcal{L}_{\text{shape}} & Waveform-shape loss (mean of $\text{SRS}_{\text{shape}}$ over $B$ and $F$); see Eq.~\eqref{eq:lshape}. \\
\mathcal{L}_{\text{TS}} & Time-series MSE between $\mathbf{x}$ and $\hat{\mathbf{x}}$; see Eq.~\eqref{eq:lts}. \\
\mathcal{L}_{\text{PSD}} & PSD MSLE Loss; see Eq.~\eqref{eq:lpsd}.\\
\mathcal{L}_{\text{SRS}} & SRS Loss: MSE between $\log \mathrm{SRS}(\mathbf{x})$ and $\log \mathrm{SRS}(\hat{\mathbf{x}})$; see Eq.~\eqref{eq:lsrs}. \\
\mathcal{L}_{\text{total}} & Total training loss; weighted sum of all terms; see Eq.~\eqref{eq:ltotal}. \\
\lambda_{\text{TS}},\,\lambda_{\text{SRS}},\,\lambda_{\text{KL}},\,\lambda_{\text{shape}},\, \lambda_{\text{PSD}} & Weights for time-series, spectral, KL, Shape, and PSD Loss terms in $\mathcal{L}_{\text{total}}$. \\
\bottomrule
\end{tabularx}
\caption{Notation: Machine learning variables for VAE and CVAE models.}
\label{tab:notation_ml}
\end{table}

\end{document}